\documentclass[10pt,twocolumn,letterpaper]{article}
\usepackage{cvpr}      
\usepackage[dvipsnames]{xcolor}
\newcommand{\red}[1]{{\color{red}#1}}

\usepackage{graphicx}
\usepackage{amsmath}
\usepackage{amssymb}
\usepackage{stmaryrd}
\usepackage{booktabs}
\usepackage{algorithm2e, algorithmic}
\usepackage{wrapfig}

\RestyleAlgo{ruled}
\SetKwComment{Comment}{/* }{ */}

\usepackage{multirow}

\usepackage{pgfplots}
\usepackage{tikz}
\pgfplotsset{compat=1.11,
    /pgfplots/ybar legend/.style={
    /pgfplots/legend image code/.code={%
       \draw[##1,/tikz/.cd,yshift=-0.25em]
        (0cm,0cm) rectangle (3pt,0.8em);},
   },
}

\newcommand{\keypoint}[1]{\vspace{0.1cm}\noindent\textbf{#1}}
\usepackage{makecell}
\usepackage{multirow}
\newcommand{\cut}[1]{}
\definecolor{cvprblue}{rgb}{0.21,0.49,0.74}
\usepackage[pagebackref,breaklinks,colorlinks,citecolor=cvprblue]{hyperref}

\usepackage[capitalize]{cleveref}
\crefname{section}{Sec.}{Secs.}
\Crefname{section}{Section}{Sections}
\Crefname{table}{Table}{Tables}
\crefname{table}{Tab.}{Tabs.}

\def\confName{CVPR}
\def\confYear{2024}

\title{What Sketch Explainability \textit{Really} Means for Downstream Tasks \vspace{-0.2cm}}

\author{\href{https://hmrishavbandy.github.io/}{Hmrishav Bandyopadhyay}\textsuperscript{1} \hspace{.2cm} 
\href{http://www.pinakinathc.me/}{Pinaki Nath Chowdhury}\textsuperscript{1} \hspace{.2cm} 
\href{https://ayankumarbhunia.github.io/}{Ayan Kumar Bhunia}\textsuperscript{1} \\ \href{https://aneeshan95.github.io/}{Aneeshan Sain}\textsuperscript{1} \hspace{.2cm}
\href{https://www.surrey.ac.uk/people/tao-xiang}{Tao Xiang}\textsuperscript{1,2} \hspace{.2cm} \href{https://personalpages.surrey.ac.uk/y.song/}{Yi-Zhe Song}\textsuperscript{1,2} \\
\textsuperscript{1}SketchX, CVSSP, University of Surrey, United Kingdom.  \\
\textsuperscript{2}iFlyTek-Surrey Joint Research Centre on Artificial Intelligence.\\
{\tt\small \{h.bandyopadhyay, p.chowdhury, a.bhunia, a.sain, t.xiang, y.song\}@surrey.ac.uk}\vspace{-0.3cm}}

\begin{document}
\maketitle

\begin{abstract}
In this paper, we explore the unique modality of sketch for explainability, emphasising the profound impact of human strokes compared to conventional pixel-oriented studies. Beyond explanations of network behavior, we discern the genuine implications of explainability across diverse downstream sketch-related tasks. We propose a lightweight and portable explainability solution -- a seamless plugin that integrates effortlessly with any pre-trained model, eliminating the need for re-training. Demonstrating its adaptability, we present four applications: highly studied retrieval and generation, and completely novel assisted drawing and sketch adversarial attacks. The centrepiece to our solution is a stroke-level attribution map that takes different forms when linked with downstream tasks. By addressing the inherent non-differentiability of rasterisation, we enable explanations at both coarse stroke level (SLA) and partial stroke level (P-SLA), each with its advantages for specific downstream tasks.
\vspace{-4mm}
\end{abstract}

\section{Introduction}
\label{sec:intro}

\begin{figure}
    \centering
    \includegraphics[trim=0.3cm 0cm 0cm 0cm, width=\linewidth]{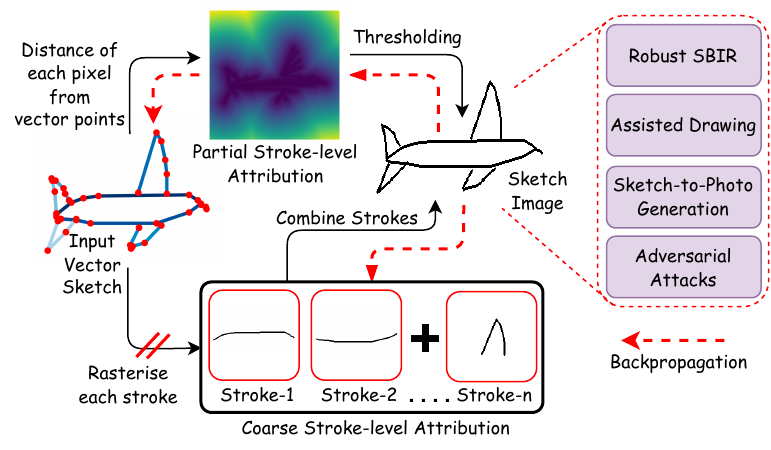}
    \caption{We attribute explanations for individual strokes (stroke-level attribution) and their vector coordinate points (point-level attribution). Stroke-level attribution rasterises individual strokes (non-differentiably) to produce $n$-stroke images. Next, we sum the stroke images to get the complete sketch image used for downstream tasks. Point-level Attribution computes distance transform from stroke coordinates and thresholds to get the sketch image. Our explainability solution works without re-training for existing tasks like SBIR and sketch-to-photo generation and novel tasks like filtering noisy strokes for assisted drawing and adversarial attack by removing a small stroke.}
    \vspace{-6mm}
    \label{fig:teaser}
\end{figure}

Sketches, rooted in human expression \cite{hertzmann2020line}, offer a distinctive modality for exploring explainability \cite{sketchXAI, yongangXAI}. In contrast to photos, where each pixel is independent and lacks inherent meaning, sketches are organised into strokes, with each stroke carrying subjective meaning assigned by the sketcher \cite{holinaty2021drawing}. This paper explores sketch explainability, but with a unique perspective -- aiming to provide explanations and unravel the true implications of explainability on various downstream sketch-related tasks.

With this perspective in mind, our approach champions an explainability solution that is (i) lightweight and portable -- a plugin seamlessly integrating with multiple pre-trained models without necessitating re-training \cite{turbe2023posthoc}, and (ii) easily adaptable to a diverse array of downstream sketch-specific tasks, benefiting the broader community.

Our solution is exclusively centred on human strokes, aiming to attribute explanation on different stroke granularity: individual strokes (coarse) and their parts (fine). The output of our model markedly differs from typical saliency maps \cite{adebayo2018sanity} found in photo-based explainability models, where the emphasis is mostly on visualisation \cite{springenberg2014guided_backprop}. Ours is a task-driven attribution map that assigns \textit{stroke-level} attributes capturing how altering stroke characteristics can impact model prediction. Depending on the downstream tasks, attributions can be grounded to, for example (i) importance of entire strokes, which is more suitable to filter noisy strokes \cite{strokesubset} in assisted drawing, and remove small strokes for adversarial attacks on existing sketch encoders, and (ii) stroke shape and length, where a partial-stroke level attribution is beneficial for tasks like sketch-based image retrieval \cite{sain2023clip} and sketch-to-photo generation \cite{koley2023picture}.

To showcase the adaptability of our model, we carefully devise four applications: two well-studied tasks from existing literature (retrieval \cite{bhunia2020sketch, collomosse2019livesketch} and generation \cite{koley2023picture, sketchhairsalon}), and two entirely novel tasks (assisted drawing and sketch adversarial attack). In \textit{retrieval}, we evaluate reliability of model predictions by comparing predicted stroke order with the order in which a human draws them. For \textit{generation}, we pinpoint strokes with the least influence, offering explicit feedback to end-users regarding which strokes the model prioritised and which it overlooked. In \textit{assisted drawing} \cite{dreamsheets}, we assist novice artists in faithfully sketching a particular photo by identifying strokes that do not match the target photo. Lastly, in \textit{adversarial attacks}, we unveil the vulnerability of state-of-the-art sketch encoders by removing a small imperceptible stroke in any sketch, resulting in significant changes to the model's prediction.

The focal point connecting all downstream tasks is our proposed stroke-level attribution. The key question, therefore, is how to backpropagate information to strokes while addressing the inherent non-differentiability of rasterisation -- strokes are most often represented as discrete coordinates and rasterised before feeding into downstream applications. We provide two solutions for non-differentiability: (i) coarse stroke level: we first rasterise individual strokes to produce raster stroke images. Then, we combine these stroke images to get the complete sketch image (see \cref{fig:teaser}) -- since this addition of stroke images is a differentiable operation, we can backpropagate information from the complete sketch to individual raster stroke images. (ii) fine partial-stroke level: we create a distance transform image from stroke coordinates (``red dots'' in \cref{fig:teaser} vector sketch) by calculating the minimum distance of each pixel in the image from the coordinates. Then, we threshold the distance value to get the sketch image (white pixels for a high distance and black pixels for a low distance). Since the distance function and our threshold step are both differentiable, we can backpropagate information from the sketch images to stroke coordinates.

Our contributions can be summarised as follows: {(i)} We explore sketch explainability, emphasising the importance of strokes in human-drawn sketches. {(ii)} We highlight the profound impact of explainability on various sketch-related domains, presenting applications in retrieval, generation, assisted Drawing, and adversarial attacks. {(iii)} We solve for the non-differentiability problem of rasterisation, and provide both stroke-level and partial-stroke level attribution.

\vspace{-1mm}
\section{Related Works}
\label{sec:related-works}
\vspace{-2mm}

\keypoint{Sketch for Visual Understanding:}
Having a high visual proximity \cite{hertzmann2020line} to real images and carrying human subjectivity \cite{sain2021stylemeup, bandyopadhyay2024INR}, amateur sketches or abstract line drawings~\cite{li2019photo} has been a popular modality for customised expression, thus driving extensive applications as a query for retrieval \cite{dutta2019semantically, sain2021stylemeup,collomosse2019livesketch, shen2018zero} of object \cite{dey2019doodle} or scene \cite{fscoco} images, 3D shapes \cite{xu2022domain}, and even concepts like in `pictionary-like' games \cite{bhunia2020pixelor}. As a canvas for creativity, sketch helps image-editing \cite{zeng2022sketchedit, liu2021deflocnet, yu2019free}, or generation of objects \cite{chen2018sketchygan, gao2020sketchycoco, wang2021sketch, chen2020deepfacedrawing}, scenes \cite{yi2022animating}, and 3D shapes \cite{guillard2021sketch2mesh, zhang2021sketch2model, yan2020interactive, bandyopadhyay2024doodle}. Being easily editable, sketch enables interactive access to AI systems like image-segmentation \cite{zou2018sketchyscene,hu2020sketch}, object localisation \cite{tripathi2020sketch}, image-inpainting \cite{zeng2022sketchedit,yu2019free}, and incremental learning \cite{bhunia2022doodle}.  Being application-specific however, such works largely ignored \textit{explaining} the `\textit{how}' of sketch-correspondence. 
The few who did, customised training pipelines \cite{sketchXAI, yongangXAI} for niche tasks. In this work, we thus make the first attempt at visualising salient sketch-regions (strokes), as an explainability-tool (like GradCAM \cite{gradCAM} in photos) for existing pre-trained sketch-based downstream networks.

\keypoint{Explaining CNN Predictions:} CNN explanations visually highlight regions of input having the maximum influence on a model's predictions \cite{simonyan2013}. This visualisation of `\textit{salient}' regions is either through an analysis \cite{fong2017interpretable} of a regular pre-trained network after it has completed training (\textit{post-hoc}), or by designing and training explicitly interpretable (\textit{i.e. explain-and-predict}) models \cite{brendel2019approximating, chen2019looks, nauta2021neural}. Given a pre-trained CNN, a \textit{post-hoc} algorithm either visualises (i) \textit{model attributes} like feature and activation-maps \cite{olah2017feature, erhan2009visualizing, nguyen2016multifaceted, zeiler2014visualising, zhou2016learning, selvaraju2017grad} that \textit{imply} saliency of specific input (pixel) regions, or (ii) \textit{input attributes} directly as pixels \cite{tobias2015simplicity} or pixel-regions \cite{kapishnikov2019xrai} coloured according to their relative importance. Visualisation of input attributes is facilitated through perturbation based algorithms \cite{fong2017interpretable,fong2019understanding, cao2015look} and gradient-based analysis \cite{simonyan2013, smilkov2017smoothgrad, tobias2015simplicity}. Perturbation-based algorithms \cite{dabkowski2017real,  petsiuk2018rise, zeiler2014visualising} detect saliency of pixel-regions by measuring impact of their absence on the prediction score. Whereas, gradient-based algorithms \cite{bach2015lrp, shrikumar2017deeplift, sundararajan2017axiomatic, dhamdhere2018important, zhang2018top} measure the gradient of the prediction with respect to individual input-pixels (input-gradients \cite{shah2021input}), attributing them based on this value. Unlike images however, sketch is a sparse-information modality \cite{chowdhury2022partially} for pixels. As such, we explore explainability in sketches by \cut{grouping sketch-pixels as strokes and }computing stroke and point attribution for fine-grained sketch explanations.

\keypoint{Evaluation of CNN Explanations:} The evaluation of CNN explanations has evolved over time -- from naive qualitative analysis of visualisations \cite{selvaraju2017grad, zeiler2014visualising, simonyan2013, tobias2015simplicity} to standardised theoretical \cite{sundararajan2017axiomatic, ancona2017towards} and empirical \cite{yeh2019fidelity, zeiler2014visualising, rao2022towards} baselines. Analysing explanations theoretically helps evaluate them in a model-agnostic environment, where their mathematical form is checked against pre-defined properties \textit{(axioms)}\cite{kindermans2019reliability, ancona2017towards, sundararajan2017axiomatic}. Empirical evaluations, on the other hand, involve experiments measuring \textit{(i)} variance in explanations upon perturbations of inputs \cite{yeh2019fidelity} and model weights \cite{adebayo2018sanity, adebayo2020debugging} (sanity checks), and \textit{(ii)} accuracy of explanations in locating important features \cite{fong2017interpretable, zhang2018top, selvaraju2017grad, rao2022towards}. These features, when perturbed, influence the CNN maximally, as measured by perturbation-based metrics \cite{kapishnikov2019xrai,petsiuk2018rise, zeiler2014visualising, selvaraju2017grad, samek2016evaluating, ancona2017unified, shah2021input}. Recently, however, these evaluation protocols have met criticism \cite{fel2021cannot} due to their detachment from humans. Instead, human studies \cite{kim2022hive, selvaraju2017grad, lage2019evaluation, sixt2022users}, and human-centered metrics \cite{fel2021cannot} have been proposed for evaluating explanation interpretability. In this work, we attribute strokes by backpropagation \cite{sundararajan2017axiomatic, tobias2015simplicity, simonyan2013,shrikumar2017deeplift}, evaluating attributions through sanity checks \cite{adebayo2018sanity}, empirical metrics \cite{kapishnikov2019xrai} and human studies \cite{kim2022hive} on downstream sketch-based applications \cite{yu2016sketch}.

\vspace{-1mm}
\section{Background}
\label{sec:background}
\vspace{-2mm}
Here, we provide a brief overview of some standard concepts, ubiquitous in explainability literature \cite{adebayo2018sanity} to help formalise the question: \textit{``what entails a good explanation?''}

\keypoint{Attribution Algorithms:} It highlights relevant regions (e.g., pixels in an image, see \cref{fig:teaser}) that are responsible for the model's prediction. Despite its importance for safety-critical applications \cite{wei2022interpretablesafety, jia2023bayesian, rudin2019blackbox}, making an attribution algorithm interpretable to humans remains an open problem. Surprisingly, a more faithful attribution is usually less interpretable and vice-versa \cite{zeiler2014visualising, tobias2015simplicity}. Prior works \cite{gradCAM, B-cos-networks} study this trade-off between faithfulness vs. interpretability as an answer to: \textit{``What makes a good visual explanation?''}.

The attribution map $\mathbb{A} \in \mathbb{R}^{H \times W}$ is typically calculated using gradients for an input $\mathrm{X} \in \mathbb{R}^{H \times W \times 3}$ for a classification model $\hat{y} = F_{\theta}(\mathrm{X}) \in \mathbb{R}^{C}$, pre-trained on $C$ categories. The gradients are a simple and good indicator of how much the model prediction changes for input $\mathrm{X}$ as,
\vspace{-1mm}
\begin{equation}\label{eq:background-attribution}
    \mathbb{A} = \partial F_{\theta} (\mathrm{X}) / \partial \mathrm{X}
    \vspace{-2mm}
\end{equation}

\keypoint{Interpretability:} It is the ability of an attribution algorithm to provide a qualitative ``understanding'' for a model \cite{LIME}. This ``understanding'' depends on the target audience, e.g., a human expert may interpret a small Bayesian network \cite{kulesza2015bayesian}, but a layman is more comfortable with a weighted attention (or feature) map that highlights salient regions \cite{adebayo2018sanity}. To evaluate the interpretability of attribution maps, prior works either (i) perform downstream tasks \cite{gradCAM, dominici2023sharcs} that depend on the interpretation (e.g., object localisation -- predict the bounding box and semantic segmentation for image region with the highest attribution) or (ii) human studies \cite{kim2022hive}, typically conducted in two setups -- class discriminative (given an attribution map, ask users to identify the category predicted by a model), and trustworthiness (compare attribution maps from a strong and a weak model, and ask users to identify the stronger model).

\keypoint{Faithfulness:} It is the ability of attribution algorithms to accurately ``explain'' the computation learned by a model. For example, in theory, a fully faithful attribution is the entire model (e.g., ResNet-18 \cite{HeResNet}) but is not interpretable by a human. In practice, for an attribution to be meaningful, it is often impossible to be completely faithful. To balance this trade-off, prior works explore human-interpretable attributions that are locally faithful for a given model prediction. One approach is image occlusion \cite{LIME}, where the difference in the model scores is measured when masking different patches in an input image. Image patches that significantly change the model score are deemed important by the attribution algorithm.

\section{Proposed Method}
\label{sec:methodology}
The attribution map $\mathbb{A}$ in \cref{eq:background-attribution} gives a faithful fine-grained explanation for each pixel in $\mathrm{X}$. However, such pixel attribution does not provide meaningful explanations when interpreting sparse human-drawn sketches (many empty white pixels). Additionally, pixel attribution does not consider the \textit{sketch construction process} -- humans sketch a sequence of strokes, not pixels. Hence, it is most appropriate that sketch predictions should be attributed to strokes and not pixels. However, the key challenge to stroke attributions is that most sketch applications \cite{sain2023clip, controlnet} use raster sketches -- a non-differentiable process to convert a sequence of strokes into pixels. In the following sections, we propose two stroke attribution algorithms that consider the sketch construction process by designing a differentiable rasterisation pipeline for a vector sequence of strokes.
\vspace{-0.2cm}

\begin{algorithm}
    \caption{Non-differentiable Rasterisation}\label{alg:naive-rasterisation}
    \KwData{$\mathrm{V} \gets$ Vector Sketch of size $\mathbb{R}^{T \times 5}$}
    \KwResult{$\mathrm{X} \gets $ Blank (Zero) Canvas of size $\mathbb{R}^{h \times w \times 3}$ \;}
    $\mathbb{B}(\cdot, \cdot) \gets$ Bresenham Function \;
    $v_{0} = (x_{0}, y_{0}, q_{0}^{1}, q_{0}^{2}, q_{0}^{3}) \gets \mathrm{V}[0]$ \;
    $v_{\text{prev}} \gets v_0$ \;
    $q_{\text{prev}} \gets q^{1}_0$ \;
    \For{$v_{t} = (x_{t}, y_{t}, q_{t}^{1}, q_{t}^{2}, q_{t}^{3}) \gets \mathrm{V}[1 \dots T]$}{
        \If{ $q_{\textnormal{prev}} = 1$ and $q_{t}^{1} = 1$}{
            \For{$(p_{x}^{i}, p_{y}^{i}) \gets \mathbb{B}(v_{\textnormal{prev}}, v_{t})$}{
                $\mathrm{X} \ ( p_{x}^{i}, p_{y}^{i} ) \gets 255$\;
            }
        }
        $v_{\text{prev}} \gets v_{t}$\;
        $q_{\text{prev}} \gets q_{t}^{1}$\;
        
        \If{$q_{t}^{3} = 1$}{ 
            $\text{exit}()$ \Comment*[r]{End of Drawing}
        }
    }
\end{algorithm}

\vspace{-0.5cm}
\subsection{Sketch Representations}
While sketches are used in several formats (or representations) like Raster \cite{yu2015sketch}, Vector \cite{ha2017neural}, or B\'ezier \cite{das2020beziersketch}, digital sketches are primarily captured in vector form, as a list of points traced on a drawing pad. These points are usually a five-element vector $v_{t} = (x_{t}, y_{t}, q_{t}^{1}, q_{t}^{2}, q_{t}^{3})$ where, $(x_{t}, y_{t})$ are absolute coordinates in a $(H \times W)$ drawing canvas and the last three elements are one-hot encoding of pen-states: pen touching the paper $(1, 0, 0)$, pen is lifted $(0, 1, 0)$, and end of drawing $(0, 0, 1)$. Hence, a vector sketch with $T$ points is represented as $\mathrm{V} \in \mathbb{R}^{T \times 5}$.

As most downstream sketch applications work on rasterised sketches $\mathrm{X} \in \mathbb{R}^{H \times W \times 3}$, prior works \cite{yu2015sketch} translate (non-differentiably) a vector sketch $\mathrm{V}$ to its equivalent raster sketch $\mathrm{X}$ using \cref{alg:naive-rasterisation}. For this, the Bresenham function $\mathbb{B}(\cdot)$ is used to connect two vector points $\{v_{t-1}, v_{t}\}$ $\in$ $\mathrm{V}$ in the pixel space $\{(p_{x}^{1}, p_{y}^{1}), \dots, (p_{x}^{n}, p_{y}^{n})\}$ $\in$ $\mathbb{B}(v_{t-1}, v_{t})$ via a continuous line. Next, we show how our sketch attributions overcome this non-differentiable rasterisation and backpropagate gradients (\cref{eq:background-attribution}) from pixels $\mathrm{X} \in \mathbb{R}^{H \times W \times 3}$ to strokes and points in $\mathrm{V} \in \mathbb{R}^{T \times 5}$.

\begin{figure}
    \centering
    \includegraphics[width=\linewidth]{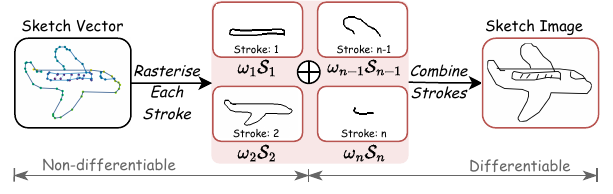}
    \vspace{-8mm}
    \caption{Coarse Stroke-level Attribution. Backpropagate gradients from raster sketch $\mathrm{X}$ to raster strokes $\mathcal{S}_{i}$, with weight $\omega_{i}$.}
    \label{fig:rsa-schema}
    \vspace{-5mm}
\end{figure}

\subsection{Coarse Stroke-level Attribution (SLA)}
In this section, we backpropagate gradients from pixel space in $\mathrm{X} \in \mathbb{R}^{H \times W \times 3}$ to strokes, defined as a continuous set of points $\{v_{t}, v_{t+1}, \dots v_{t+n}\}$ from the \textit{first} pen-down $(1, 0, 0)$ till the pen-up $(0, 1, 0)$ state. \cref{alg:naive-rasterisation} converts the vector stroke points into a raster stroke $\mathcal{S}_{i} \in \mathbb{R}^{H \times W \times 3}$. The final raster sketch is then a differentiable composition\footnote{For overlapping strokes in $\mathcal{S}_{i}$ and $\mathcal{S}_{j}$, we clamp the maximum pixel value using differentiable functions like  \texttt{torch.clamp}} of $m$ raster strokes $\mathrm{X} = \sum^m_{k=1} \mathcal{S}_{k}$. We compute SLA as,
\begin{equation}\label{eq: rsa-attribution}
    \begin{split}
        \mathbb{A}_{i}^{R} = \frac{\partial F_{\theta}(\mathrm{X})}{\partial \mathcal{S}_{i}} = \frac{\partial F_{\theta}(\mathrm{X})}{\partial \mathrm{X}} \cdot \frac{\partial \sum^m_{k=1} \mathcal{S}_{k}}{\partial \mathcal{S}_{i}}
    \end{split}
\end{equation}
This, however, gives a degenerate solution where all strokes will have the same attribution $\partial \sum^m_{k=1} \mathcal{S}_{k} / \partial (\mathcal{S}_{i}) = 1$. To avoid this, we compute a weight factor $\omega_{i} \in \mathbb{R}^{H \times W \times 3}$ for each stroke $\mathcal{S}_{i}$ such that,
\begin{equation}\label{eq: rsa-weight}
    \omega_{i} (p_{x}, p_{y}) = \begin{cases}
        1 & \text{ if } (p_{x}, p_{y}) \in \mathbb{B}(v_{t-1}, v_{t}) \\
        0 & \text{ otherwise}
    \end{cases}
\end{equation}
In other words, given consecutive vector sequence of points $(v_{t-1}, v_{t})$ in stroke $\mathcal{S}_{i}$, we find all points $(p_{x}, p_{y})$ using Bresenham function $\mathbb{B}(\cdot)$ that lie ``on the stroke'' and assign $\omega_{i}(p_{x}, p_{y})=1$. Hence, longer strokes will have more $1$'s compared to shorter strokes. Finally, \cref{eq: rsa-attribution} is adapted as 
\begin{equation}\label{eq: rsa-attribution-weight}
    \begin{split}
        \mathbb{A}_{i}^{R} = \frac{\partial F_{\theta}(\mathrm{X})}{\partial \mathrm{X}} \cdot \frac{\partial \sum_{k=1}^m\omega_{k} \mathcal{S}_{k}}{\partial \mathcal{S}_{i}}
    \end{split}
\end{equation}
SLA makes the non-differentiable rasterisation ($\mathrm{V}$$\rightarrow$$\mathrm{X}$) \textit{partially differentiable} (stokes $\mathcal{S}$ to sketch $\mathrm{X}$). In other words, SLA answers \textit{``Which strokes in a sketch are important''}. Next, we make the rasterisation fully differentiable and backpropagate gradients to vector points $\mathrm{V}$ to answer \textit{``Which point in a sketch is important''}.

\begin{figure}
    \centering
    \includegraphics[width=0.92\linewidth]{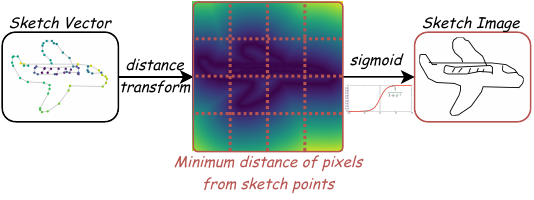}
    \vspace{-4mm}
    \caption{Partial Stroke-level Attribution. Backpropagate gradients from raster sketch $\mathrm{X}$ to vector sequence of coordinates $\mathrm{V}$.}
    \label{fig:vsa-schema}
    \vspace{-6mm}
\end{figure}

\subsection{Partial Stroke-level Attribution (P-SLA)}
Unlike SLA, which partially captures the sketch construction process (stroke-level), partial stroke-level attribution (P-SLA) can fully backpropagate gradients to the vector list of coordinates $\mathrm{V} \in \mathbb{R}^{T \times 5}$ traced on a drawing pad. Given a blank canvas $\mathrm{X} \in \mathbb{R}^{H \times W \times 3}$, we (i) calculate the minimum distance of every pixel ($p_{x}, p_{y}$) in $\mathrm{X}$ from a line segment ($v_{t-1}, v_{t}$) in $\mathrm{V}$, and (ii) compute the pixel intensity of $\mathrm{X}(p_{x}, p_{y})$ as function of the minimum distance as
\begin{equation}\label{eq:vsa-pixel}
    \begin{split}
        \mathrm{X}(p_{x}, p_{y}) = & \sigma \big[ 2 - 5 \min_{t=2, \dots, T} \big(\\
        & \hspace{-1cm} \texttt{dist}((p_{x}, p_{y}), v_{t-1}, v_{t}) + (1 - q_{t-1}^{1}) 10^{6} \big) \big]
    \end{split}
\end{equation}
where $\sigma(\cdot)$ is the sigmoid function, and $\texttt{dist}(\cdot)$ is a distance function (see Supp.) from a point ($p_{x}, p_{y}$) to a line segment ($v_{t-1}, v_{t}$). For pen-up states ($q_{t-1}^{1}=0$), we blow up ($\times 10^{6}$) the distance values that make the pixel intensities $\mathrm{X}(p_{x}, p_{y}) \to 0$, i.e., not render strokes for ($v_{t-1}, v_{t}$). Finally, we compute P-SLA
\begin{equation}\label{eq: vsa}
    \mathbb{A}_{t}^{V} = \frac{\partial F_{\theta}(\mathrm{X})}{\partial v_{t}} = \frac{\partial F_{\theta}(\mathrm{X})}{\partial \mathrm{X}} \cdot \sum_{\forall p_{x}, p_{y}} \Big\{ \frac{\partial \mathrm{X}(p_{x}, p_{y})}{\partial v_{t}} \Big\}
\end{equation}

\section{Applications of Stroke Attributions} \label{sec: applications}
Despite its simplicity, designing stroke attribution algorithms that capture the sketch construction process unlocks insights into numerous existing downstream tasks like classification \cite{ha2017neural}, robust sketch-based image retrieval (category-level \cite{berlin} and fine-grained \cite{sangkloy2016sketchy}) and enables some novel sketch applications like assisted drawing \cite{strokesubset}, interactive sketch to photo generation \cite{koley2023picture}, adversarial attacks on human-drawn sketches, and discovering the ``arrow of time'' \cite{arrow-of-time} in raster sketches.

\subsection{Robust Sketch Based Image Retrieval}\label{sec: retrieval}
Given a query sketch $\mathrm{X} \in \mathbb{R}^{H \times W \times 3}$, category-level sketch-based image retrieval (SBIR) aims to fetch category-specific photos from a gallery of multi-category photos (\eg, given sketch of a `shoe' retrieve \textit{any} photo `shoe' from a gallery of `shoes+hats+cows'). Conversely, fine-grained SBIR aims to retrieve \textit{one} instance from a gallery of the \textit{same} category photos (e.g., given the sketch of a `shoe' retrieve \textit{one} photo shoe from a gallery of \textit{all} shoes). Deep learning frameworks learn a joint sketch-photo manifold (for category and fine-grained) via a feature extractor \cite{collomosse2019livesketch, dey2019doodle, xu2018sketchmate} trained using triplet loss \cite{yu2016sketch}. Recent adoption of foundation models for SBIR \cite{sain2023clip} shifts focus to robust deployment using the open-set generalisation of CLIP \cite{CLIP}.

\begin{figure}[t]
    \centering
    \includegraphics[width=\linewidth]{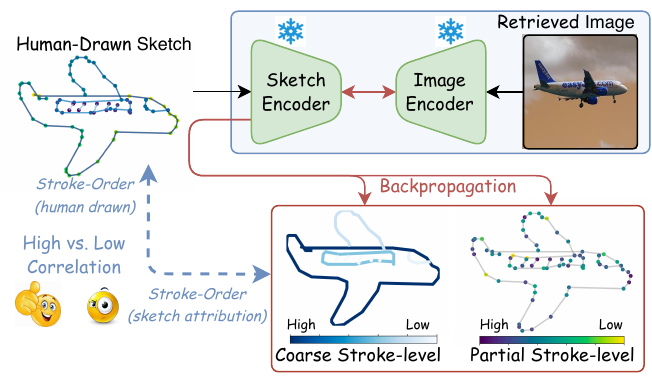}
    \vspace{-5mm}
    \caption{Sketch attributions from stroke-level and point-level for image retrieval. High correlation of human-drawn stroke order with that from sketch-attributions (high$\to$low) indicate our sketch encoder gives more importance to salient strokes drawn \textit{early on}.}
    \label{fig:retrieval-SLA-P-SLA}
    \vspace{-4mm}
\end{figure}

Towards this goal of robust deployment, our sketch attribution algorithms ($\mathbb{A}^{R}_{i}$ and $\mathbb{A}^{V}_{t}$) can predict which strokes the network focuses on when retrieving a photo (\cref{fig:retrieval-SLA-P-SLA}). Apart from interpreting SBIR models, sketch attribution can also help detect potential failures at runtime (inference). First, we use the attribution scores $\mathbb{A}^{R}_{i}$ or $\mathbb{A}^{V}_{t}$ to rearrange the strokes from highest to lowest. The attribution scores indicate the most salient to the least salient strokes that affect model prediction. Second, we calculate a correlation ($Corr$) of our predictor stroke order with the ground-truth temporal stroke order drawn by a user (humans draw the most salient regions first and least salient areas last \cite{berlin, sangkloy2016sketchy}). A high correlation indicates that humans and our model prioritise strokes similarly, whereas a low $Corr$ denotes that the model and the user prioritise different strokes. We evaluate SBIR on a pre-trained SOTA model \cite{sain2023clip} using CLIP with prompt learning as a sketch and photo encoder.

\keypoint{Datasets:} We use TU-Berlin \cite{berlin} (for category-level SBIR) and Sketchy \cite{sangkloy2016sketchy} (for fine-grained SBIR). TU-Berlin contains $250$ categories, with $80$ free-hand sketches in each, and $204,489$ images \cite{zhang2016sketchnet}. Sketchy \cite{sangkloy2016sketchy} has $75,471$ sketches over $125$ categories having $100$ images in each \cite{yelamarthi2018zero}.

\keypoint{Evaluation Metrics:} Following \cite{sain2023clip}, we use mean average precision (mAP) and precision for top $200$ retrieved samples (P@200) for category-level SBIR. For fine-grained SBIR \cite{chowdhury2022partially}, we measure Acc.@q, i.e., the percentage of sketches whose true matched photo is in the top-q list.

\keypoint{Results:} We divide the evaluation set into two sets: (i) those that have a high correlation $Corr \geq 0.5$, and (ii) those with a low correlation $Corr \leq 0.1$ of ground-truth and predicted stroke order. \cref{tab:attribution-correlation-retrieval} shows sketches with a high $Corr \geq 0.5$ are $1.7/3.9$ more accurate in Acc.@1/Acc.@5 than those with $Corr \leq 0.1$ for fine-grained SBIR, and $3.3/1.7$ better in mAP/P@200 for category-level SBIR. \textbf{Full Dataset} indicates the performance of the pre-trained model on the entire evaluation set.

\begin{table}
    \centering
    \scriptsize
    \caption{Stroke attribution (SLA, P-SLA) make SBIR systems reliable. Sketches with a high correlation ($Corr$) of stroke saliency (predicted by SLA or P-SLA) with human-drawn temporal stroke order tend to have higher retrieval accuracy.}
    \begin{tabular}{ccccccc}
        \toprule
          & \multirow{2}{*}{\textbf{Metrics}} & \multirow{2}{*}{\textbf{Full Dataset}} & \multicolumn{2}{c}{$\mathbf{Corr \geq 0.5}$} & \multicolumn{2}{c}{$\mathbf{Corr \leq 0.1}$} \\
          & &                  & \textbf{SLA} & \textbf{P-SLA} & \textbf{SLA} & \textbf{P-SLA} \\ \midrule
        \multirow{2}{*}{\makecell{Category\\Level}} & mAP & 53.1 & 55.3 & 57.6 & 51.7 & 50.1 \\ 
         & P@200 & 65.9 & 66.7 & 68.5 & 64.6 & 61.5 \\ \midrule
         \multirow{2}{*}{\makecell{Fine\\Grained}} & Acc.@1 & 15.3 & 16.4 & 17.6 & 13.8 & 12.7 \\ 
         & Acc.@5 & 34.2 & 36.9 & 39.4 & 31.1 & 28.3 \\ \bottomrule
    \end{tabular}
    \label{tab:attribution-correlation-retrieval}
    \vspace{-5mm}
\end{table}

\vspace{-1mm}
\subsection{Assisted Drawing via Noisy Stroke Removal}
\vspace{-2mm}

Although sketching has enabled many exciting applications \cite{controlnet, koley2023picture, chowdhury20223Dsynthesis, t2i-adapter}, the fear to sketch (i.e., \textit{``I can't sketch''}) has proven fatal for its widespread adoption. To solve this, prior works \cite{strokesubset} used complex (and hard to train \cite{bhunia2020sketch}) reinforcement learning \cite{PPO} to predict the importance of each stroke in a sketch. Next, a stroke subset selector removes noisy (less important) strokes, leaving only those positively contributing to the downstream tasks. 

In this section, we focus on assisted drawing -- given a photo, we use attribution scores ($\mathbb{A}^{R}_{i}$ or $\mathbb{A}^{V}_{t}$) to help humans draw a faithful and clean sketch. Our method is significantly simpler than reinforcement learning alternatives \cite{strokesubset}.

For SLA, an input sketch $\mathrm{X} = \sum_{k=1}^m\omega_{k} \mathcal{S}_{k}$ is composed of $m$ strokes. We calculate the cosine similarity ($\texttt{sim}$) of input sketch $\mathrm{X}$ with its target photo $\mathrm{P} \in \mathbb{R}^{H \times W \times 3}$ to measure \textit{``how faithfully $\mathrm{X}$ describes $\mathrm{P}$''}. Next, we backpropagate gradients from the cosine similarity to strokes $\mathcal{S}_{i}$ and calculate stroke-level attribution score $\mathbb{A}^{R}_{i}$ as
\begin{equation}
    \mathbb{A}_{i}^{R} = \frac{\partial \texttt{sim}( F_{\theta}(\mathrm{X}), F_{\theta}(\mathrm{P}) )}{\partial \mathrm{X}} \cdot \frac{\partial \sum^m_{k=1}\omega_{k} \mathcal{S}_{k}}{\partial \mathcal{S}_{i}}
\end{equation}
The pre-trained sketch and photo encoder\footnote{The sketch and photo encoder $F_{\theta}(\cdot)$ could be a siamese-style shared network or two independent models with different network weights \cite{sain2023clip}.} $F_{\theta}(\cdot)$ must be highly accurate to judge sketch--photo correspondence. Hence, we use pre-trained CLIP+prompts encoder from \cite{sain2023clip}. To remove noisy strokes, we only update the weights $\omega_{i} \in \mathbb{R}^{H \times W \times 3}$ using a normalised attribution score $\mathbb{A}^{R}_{i}$ as
\begin{equation}
    \omega_{i}^{*} = \omega_{i} \cdot \texttt{Gumbel\_Softmax}\Big(\frac{\mathbb{A}_{i}^{R}}{\sum_{\forall i} \mathbb{A}_{i}^{R}} + \Delta \Big)
\end{equation}
$\texttt{Gumbel\_Softmax}(\cdot)$ makes the output one-hot (discrete value) in the forward pass but differentiable with a probability distribution that sum to $1$ in the backward pass \cite{jang2016categorical}. The modified sketch is constructed as $\mathrm{X} = \omega_{1}^{*}\mathcal{S}_{1} + \dots + \omega_{m}^{*}\mathcal{S}_{m}$. Intuitively, we keep strokes that contribute ($\mathbb{A}^{R}_{i}$) to a high cosine similarity matching human sketch $\mathrm{X}$ and target photo $\mathrm{P}$ and remove $\mathcal{S}_{i}$ with {normalised} $\mathbb{A}^{R}_{i}$ lower than ($0.5 - \Delta$).

Similar to SLA, we can also use P-SLA to compute the attribution $\mathbb{A}^{V}_{t}$ for each point $v_{t}$ in the vector sketch $\mathrm{V} \in \mathbb{R}^{H \times W \times 3}$, by measuring the cosine similarity between input $\mathrm{X}$ and target photo $\mathrm{P}$ as
\begin{equation}
        \mathbb{A}_{t}^{V} = \frac{\partial \texttt{sim}(F_{\theta}(\mathrm{X}), F_{\theta}(\mathrm{I}))}{\partial \mathrm{X}} \cdot \sum_{\forall p_{x}, p_{y}} \Big\{ \frac{\partial \mathrm{X}(p_{x}, p_{y})}{\partial v_{t}} \Big\}
\end{equation}
We remove noisy points $v_{t}$ by updating the pen-states in \cref{eq:vsa-pixel} from pen-down ($1, 0, 0$) to pen-up ($0, 1, 0$) depending on its attribution $\mathbb{A}^{V}_{t}$ for point $v_{t} \in \mathrm{V}$.
\begin{equation}
        q_{t-1}^{1*} = q^{1}_{t-1} \cdot \texttt{Gumbel\_Softmax} \Big( \frac{ \mathbb{A}^{V}_{t-1} }{\sum_{\forall t} \mathbb{A}_{t}^{V}} \Big) \\
\end{equation}
Using updated values for $q_{t-1}^{1*}$, we update $q_{t-1}^{2*} = 1 - q_{t-1}^{1*}$ and recalculate pixel intensities for raster sketch $\mathrm{X}(p_{x}, p_{y})$. The value of hyper-parameter $\Delta$ significantly affects the stroke removal process. We found the optimal $\Delta$ for SLA and P-SLA is $0.3$ and $0.1$, respectively. A higher $\Delta$ for P-SLA gives broken lines with a drop in the visual quality of an input sketch. Next, we evaluate stroke filtering using SLA and P-SLA on popular human-drawn sketch datasets.

\begin{figure}
    \centering
    \includegraphics[width=0.8\linewidth]{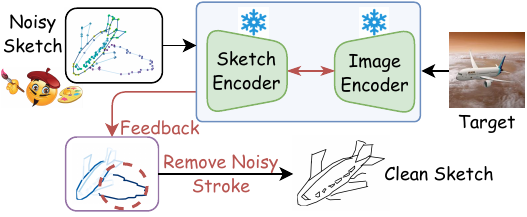}
    \caption{Assisted drawing via sketch healing (or filtering noisy strokes) using stroke attributions from SLA and P-SLA. This helps users having fear-to-sketch (\textit{``I can't sketch''}).}
    \label{fig: sketch-healing}
    \vspace{-5mm}
\end{figure}

\keypoint{Dataset:} For a fair comparison with prior works \cite{strokesubset}, we evaluate on fine-grained SBIR datasets QMUL-Shoe-V2 and QMUL-Chair-V2 \cite{yu2016sketch, bhunia2020sketch, pang2019generalising}. It consists of $6,730/1,800$ sketches and $2,000/400$ photos from Shoe-V2/Chair-V2. We evaluate on the standard test-split of $679/525$ sketches and $200/100$ photos.

\keypoint{Evaluation Metric:} We measure the retrieval accuracy of the clean sketch with the target photo by computing Acc.@1 and Acc.@5. A high accuracy need not correspond to high visual quality to the human eye \cite{sheikh2006quality}. Hence, we conducted a small human study with $5$ participants and reported the mean opinion score (MOS) \cite{thu2010mos}; each was asked to compare two sets of $50$ sketch pairs (GT sketch vs SLA filtered) and (GT sketch vs P-SLA filtered). 

\keypoint{Results:} Removing noisy strokes from human-drawn sketches is still a new topic; hence, to the best of our knowledge, there is only one work by Bhunia \etal \cite{strokesubset}. From \cref{tab: filtering}, both Bhunia \etal \cite{strokesubset}, and ours improve fine-grained SBIR performance by $10.3\%$ and $3.1\%$, respectively. However, \cite{strokesubset} outperforms our SLA and P-SLA filtered methods by $7.6\%$ and 7.2\%, respectively. This performance gap is likely because \cite{strokesubset} trains the baseline model \cite{bhunia2020sketch} using actor-critic version of PPO \cite{PPO}, whereas our SLA and P-SLA work post-hoc \cite{turbe2023posthoc} \textit{without training} the baseline model \cite{bhunia2020sketch}. Additionally, Bhunia \etal \cite{strokesubset} aims to design a robust SBIR pipeline, whereas our SLA/P-SLA filtering aims to assist humans in drawing sketches. For human study (MOS): (i) users prefer SLA-filtered $78.5\%$ vs. $21.5\%$ for GT sketch, (ii) however, P-SLA-filtered are preferred only $57.1\%$ vs. $42.9\%$ for GT sketch. This is verified by \cref{fig: sketch-healing} where P-SLA-filtered sketches have broken strokes that degrade their visual quality.

\begin{table}
    \centering
    \scriptsize
    \caption{Noisy stroke removal using SLA and P-SLA attribution.}
    \vspace{-2mm}
    \begin{tabular}{cccccc}
        \toprule
         & \textbf{Metrics} & \textbf{\makecell{GT\\Sketch}} & \textbf{\makecell{SLA\\filtered}} & \textbf{\makecell{P-SLA\\filtered}} & \textbf{\makecell{Bhunia \etal \cite{strokesubset}}} \\ \midrule
        \multirow{3}{*}{\textbf{Shoe-V2}} & Acc.@1 & 33.4 & 36.1 & 36.5 & 43.7 \\
         & Acc.@5 & 67.8 & 68.7 & 69.3 & 74.9 \\
         & MOS & 28.6 & 85.7 & 57.1 & -- \\ \midrule
         \multirow{3}{*}{\textbf{Chair-V2}} & Acc.@1 & 53.3 & 54.9 & 56.5 & 64.8 \\
         & Acc.@5 & 74.3 & 76.6 & 77.1 & 79.1 \\
         & MOS & 35.8 & 71.4 & 57.1 & -- \\ \bottomrule
    \end{tabular}
    \label{tab: filtering}
    \vspace{-0.3cm}
\end{table}

\subsection{Interactive Sketch To Photo Generation}
The upsurge of large-scale image generation models (e.g., Stable-Diffusion \cite{stable-diffusion}, GigaGAN \cite{GigaGan}) helped develop sketch-conditional image generation \cite{controlnet, t2i-adapter}. However, a key limitation of conditional image generation is that these models do not always faithfully follow the input condition. This was resolved in two stages for text-to-image generation: (i) find word tokens with low influence on generated image, and (ii) iteratively update its activation until it reaches a minimum required value. In this section, we design a pipeline for faithful sketch-to-image generation.

Using sketch attributions from SLA and P-SLA, we design a post-hoc \cite{turbe2023posthoc} method that gives \textit{feedback to the user} -- which strokes the model \textit{focuses} on and which strokes are being \textit{ignored}. Given this feedback, a user can interact with the system to ensure the model attends all salient regions.

\begin{figure}
    \centering
    \includegraphics[width=\linewidth]{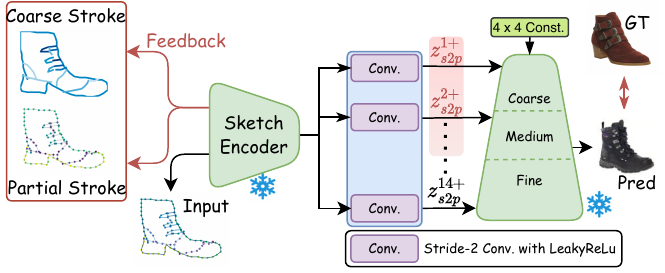}
    \caption{Interactive Sketch to Photo Generation: Our stroke attribution algorithms make existing sketch-to-photo generation pipelines \cite{koley2023picture} more \textit{faithful}. We achieve this by computing the stroke-level $\mathbb{A}^{S}_{i}$ or coordinate-level $\mathbb{A}^{V}_{t}$ attribution that has a \textit{maximal} influence on the latent code $z_{s2p}^{+}$ used by the image decoder.}
    \label{fig: interactive-generation}
    \vspace{-4mm}
\end{figure}

Our interactive pipeline is built on top of a pre-trained sketch-to-photo generation model \cite{koley2023picture}, comprising a modified ResNet-50 \cite{HeResNet} as sketch encoder and StyleGAN \cite{style-gan} as image decoder. Given a raster sketch $\mathrm{X}$, the modified sketch encoder computes a latent vector $z_{s2p}^{+} = F_{\theta}(\mathrm{X})$, where $z_{s2p}^{+} \in \mathbb{R}^{14 \times 512}$. Next, the StyleGAN \cite{style-gan} decoder generates the underlying image from $z_{s2p}^{+}$. For a faithful sketch-to-image generation, we measure the ``influence'' of each stroke/vector point on the latent code $z_{s2p}^{+}$. Particularly, prior works \cite{semantic-hierarchy} suggest that $z_{s2p}^{+}$ is disentangled into $14$-level semantic feature hierarchy, where $z_{s2p}^{+1} \in \mathbb{R}^{512}$ has coarse-level features controlling major semantic structures and $z_{s2p}^{+14} \in \mathbb{R}^{512}$ has fine-level features controlling colour schemes, etc. Since sketches primarily convey semantic structure, we use the sum of the first $7$-layers\cut{$(z_{s2p}^{+1} + \dots + z_{s2p}^{+7})$} to compute stroke attribution $\mathbb{A}^{R}_{i}$/$\mathbb{A}^{V}_{i}$ as
\begin{equation}
    \begin{split}
        \mathbb{A}_{i}^{R} = \frac{\partial \sum_{k=1}^7 z_{s2p}^{+k}}{\partial \mathrm{X}} &\cdot \frac{\partial \sum^m_{k=1} \omega_{k}\mathcal{S}_{k}}{\partial \mathcal{S}_{i}} \\
        \mathbb{A}_{t}^{V} = \frac{\partial \sum^7_{k=1} z_{s2p}^{+k}}{\partial \mathrm{X}} &\cdot \sum_{\forall p_{x}, p_{y}} \Big\{ \frac{\partial \mathrm{X}(p_{x}, p_{y})}{\partial v_{t}} \Big\}
    \end{split}
\end{equation}
Next, we qualitatively evaluate our iterative sketch to photo generation pipeline, as shown in ~\cref{fig: interactive-generation}.

\begin{figure*}
    \centering
    \includegraphics[trim=0cm 0.3cm 0cm 0cm, width=\linewidth]{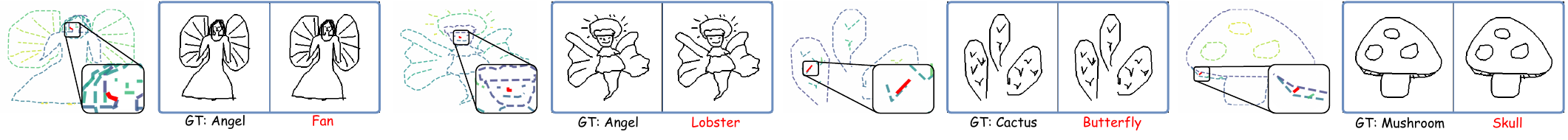}
    \caption{Adversarial attacks on human drawn sketches using SLA and P-SLA. The adversarial strokes are marked in \red{RED}.}
    \vspace{-6mm}
    \label{fig: adversarial}
\end{figure*}

\subsection{Adversarial Attacks on Human Sketches}
Szegedy \etal \cite{szegedy2013attacks} discovered that predictions by deep networks can be manipulated with extremely low-magnitude input perturbations. For images, these can be restricted to be imperceptible to human vision, but their effect can completely change the output prediction by a deep network. Such adversarial attacks are possible in image classification \cite{szegedy2013attacks}, semantic segmentation \cite{arnab2018attack, he2020attack}, object detection \cite{tu2020attack, zhang2019attack}, object tracking \cite{jia2020attack, chen2020attack}, etc. Studying these quirks is crucial as it can pose a real threat to deep learning as a pragmatic technology \cite{akhtar2021attack}. While major work has been dedicated to attacks on images, the recent surge of deployable sketch applications \cite{sain2023clip, controlnet} motivates us to \textit{present the first study} on adversarial attacks for human-sketches.

We show how our stroke attribution algorithms (SLA and P-SLA) provide the necessary information for adversarial attacks. For brevity, we focus on adversarial attacks on sketch classification \cite{yu2015sketch, ha2017neural}. Intuitively, we use sketch attribution to remove the smallest stroke (in SLA Attack) and minimum number of points (in P-SLA Attack), yet have the maximum impact on changing prediction of a pre-trained classifier $F_{\theta}(\mathrm{X}) = y_\text{cls}$. Ours is \textit{(i)} a \textit{white-box} \cite{tashiro2020diversity, phan2021adversarial} setting -- we have access to network weights and gradients of our ResNet-18 \cite{HeResNet} classifier, pre-trained on QuickDraw \cite{ha2017neural} or TU-Berlin \cite{berlin}; \textit{(ii)} \textit{untargeted} attack -- while targeted attacks \cite{zhou2020LGGAN, li2020tranferable} misclassify $F_{\theta}(\cdot)$ from $y_\text{cls}^\text{GT}$ to \textit{a specific target class} $y_\text{cls}^{*}$, untargeted attacks \cite{szegedy2013} aim to misclassify to \textit{any arbitrary class} $y_\text{cls}^\text{GT} \neq y_\text{cls}$. For a neater description of SLA and P-SLA attacks, we define the rasterisation process using $\mathcal{R}(\cdot)$ as $\mathrm{X} = \omega_{1}\mathcal{S}_{1} + \dots + \omega_{m}\mathcal{S}_{m} = \mathcal{R}(\mathcal{S})$ for SLA and following \cref{eq:vsa-pixel} for P-SLA we define $\mathrm{X} = \mathcal{R}(\{v_{1}, \dots v_{T}\}) = \mathcal{R}(\mathrm{V})$. Next, we find a stroke $\mathcal{S}_{adv}$ with stroke length $|\mathcal{S}_{i}|$ less than some threshold $\epsilon$ as
\begin{equation}
    \mathcal{S}_{adv} = \arg \max_{| \mathcal{S}_{j}| \leq \epsilon}  \mathcal{L}_\text{cls} \big( F_{\theta}( \mathcal{R}( \mathcal{S} - \{\mathcal{S}_{j}\} ) ), \ y_\text{cls}^\text{GT} \big)
\end{equation}
Unlike typical adversarial attacks on images that \textit{add} a small noise $(\mathrm{X} + \Delta x)$ with $||\Delta x||_{\infty} \leq \epsilon$, for sketch adversarial attacks, we \textit{remove} a small stroke ($\mathrm{X} - \mathcal{S}_{adv}$) such that the stroke length is less than $\epsilon$ as $|\mathcal{S}_{adv}| \leq \epsilon$. For P-SLA attack, we find a subset of $\epsilon$ vector points $\mathrm{V}_{adv} = \{v_{1}^{adv}, \dots v_{\epsilon}^{adv}\}$ from input sketch $\mathrm{V} \in \mathbb{R}^{T \times 5}$ which maximises the categorical cross-entropy loss $\mathcal{L}_\text{cls}$ as
\begin{equation}
    \begin{split}
        v_{t}^{adv} &= \mathcal{L}_\text{cls} \big( F_{\theta}( \mathcal{R}( \mathrm{V} - \{v_{t}\} )), \ y_\text{cls}^\text{GT} \big) \\
        \mathrm{V}_{adv} &= \texttt{top@k} (\{ v_{1}^{adv}, v_{2}^{adv}, \dots v_{T}^{adv} \}, \epsilon)
    \end{split}
\end{equation}
where, $\texttt{top@k}(\cdot, \epsilon)$ picks the highest $\epsilon$ elements. \cref{fig: adversarial} shows the adversarial strokes $\mathcal{S}_{adv}$ and points $\mathrm{V}_{adv}$ in \red{red}.

\keypoint{Dataset:} We evaluate sketch adversarial attacks on QuickDraw \cite{ha2017neural} and TU-Berlin \cite{berlin}. We use a subset \cite{xu2018sketchmate} of $50M$ sketches in QuickDraw \cite{ha2017neural} as $3.8M$ samples across $345$ categories split in $2.1M$ sketches for training, $0.3M$ for validation and $0.4M$ for evaluation. See \cref{sec: retrieval} for details on TU-Berlin \cite{berlin} dataset.

\keypoint{Evaluation:} We measure the drop in classification accuracy when using SLA and P-SLA attacks in \cref{tab: adversarial-attacks}. Unlike image-based adversarial attacks where $\epsilon$ is a pixel intensity (non-integer, decimal value), our sketch attacks occur on stroke/point length, making $\epsilon$ an integer. A higher $\epsilon \geq 20$ removes ``visible'' strokes in SLA and broken lines in P-SLA (\cref{fig: adversarial}). Hence, we evaluate accuracy drop for $\epsilon=5$ and $\epsilon=15$ in \cref{tab: adversarial-attacks}. We observe for $\epsilon=5$ P-SLA offers a better adversarial attack than SLA by a margin of $1.3\%$.

{
\setlength{\tabcolsep}{0.3cm}
\begin{table}
    \centering
    \scriptsize
    \caption{Sketch Adversarial Attacks: Using stroke attributions, we remove a small stroke ($|\mathcal{S}_{i}| \leq \epsilon$) that misclassifies an input sketch.}
    \vspace{-1mm}
    \begin{tabular}{cccc|ccc}
        \toprule
         & & \multicolumn{2}{c}{\textbf{SLA Attack}} & \multicolumn{2}{c}{\textbf{P-SLA Attack}} \\
         & \multirow{-2}{*}{\textbf{\makecell{No\\Attack}}} & $\epsilon=5$ & $\epsilon=15$ & $\epsilon=5$ & $\epsilon=15$ \\ \midrule
        QuickDraw & 67.2 & 65.7 & 64.5 & 65.1 & 63.7 \\
        TU-Berlin & 74.9 & 71.5 & 68.5 & 70.2 & 68.1 \\ \bottomrule
    \end{tabular}
    \vspace{-4mm}
    \label{tab: adversarial-attacks}
\end{table}
}

\vspace{-1mm}
\section{Human Study}
\vspace{-2mm}

Interpretability aims to help humans understand a model's reasoning process (\textit{transparency}), verify that its predictions are based on the right constraints (\textit{fairness}), and evaluate its confidence (\textit{trustworthiness}) \cite{gradCAM, LIME}. In \cref{sec: applications}, we evaluated interpretability using several automatic metrics (e.g., classification or retrieval accuracy) on different evaluation datasets \cite{ha2017neural, berlin, sangkloy2016sketchy}. However, highlighting salient regions by backpropagating gradients for downstream applications does not capture how helpful end-users find these attributions \cite{LIME, kim2022hive}. In this section, we take a \textit{human-centred approach to interpretability} -- how well our stroke attribution algorithms align with the reasoning process of humans and the trade-off, interpretability vs. accuracy.

\keypoint{Setup:} We recruited $7$ participants from different geographical regions, in the age group $20-30$ years. All participants had some background in AI research, but only $3$ reported having prior experience in interpretability. Once recruited, each user is assigned a unique ID for anonymity. For SLA and P-SLA, we conduct $5$ human studies, with each having $10$ multiple choice questions (MCQs). Hence, each participant answers $50$ MCQs for SLA and $50$ for P-SLA.

\begin{figure}[h]
    \centering
    \includegraphics[width=\linewidth,trim={0 2.5mm 0 6mm}]{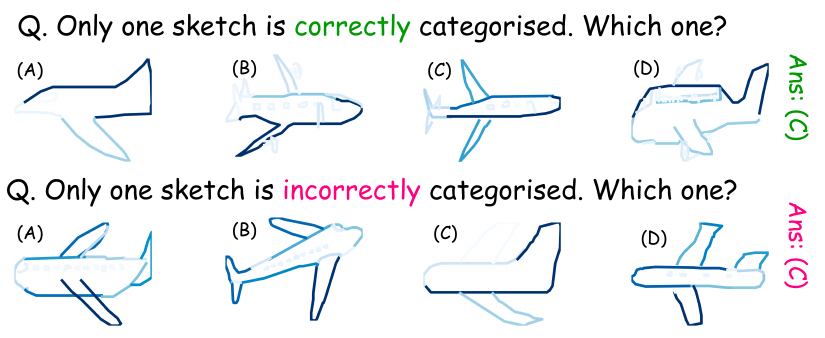}
    \caption{Evaluating transparency: Can a human understand the behaviour of an existing (pre-trained) classifier with SLA/P-SLA.}
    \vspace{-4mm}
    \label{fig: hs-transparency}
\end{figure}

\keypoint{Evaluating Transparency:}
For an attribution to be useful, humans must \textit{understand} a model's behaviour for correct and incorrect predictions. In this section, we evaluate if our SLA and P-SLA can make existing sketch classifiers (i.e., Sketch-A-Net \cite{yu2015sketch}), pre-trained on TU-Berlin \cite{berlin} dataset \textit{transparent} to humans. Accordingly, we choose a random category and select $4$ sketch instances -- (i) $3$ misclassified and $1$ correctly classified, and (ii) $1$ misclassified and $3$ correctly classified. Next, as shown in \cref{fig: hs-transparency}, we compute stroke attribution (SLA and P-SLA) for the selected sketches and ask users: ``Only $1$ of these $4$ sketches are correctly (or incorrectly) recognised by our model. Please select that correct (or incorrect) sketch.''. 
We find users can identify correct/incorrect model predictions $75.9\%$/$63.4\%$ of times for SLA and $76.3\%$/$65.2\%$ for P-SLA.

\begin{figure}[h]
    \centering
    \includegraphics[width=\linewidth,trim={0 2mm 0 6mm}]{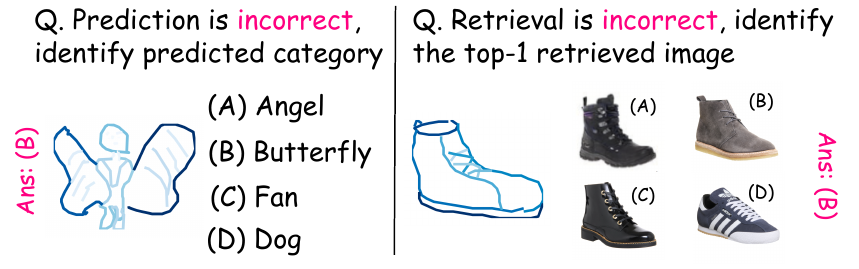}
    \caption{Evaluating Fairness: For an incorrect model prediction, we evaluate if humans can ``identify what went wrong''.}
    \vspace{-4mm}
    \label{fig: hs-fairness}
\end{figure}

\keypoint{Evaluating Fairness:} End users are much better positioned to make a decision with help from a model if intelligible explanations are provided. In this section, we evaluate if SLA and P-SLA can help end users understand \textit{``What went wrong?''} (i) For a pre-trained sketch classifier \cite{yu2015sketch}, we show users a misclassified sketch instance and ask users to identify the (wrongly) predicted category. (ii) For fine-grained SBIR \cite{bhunia2021more}, we show a sketch (whose GT photo is not in top-$10$) and ask users to identify the top-$1$ (wrongly) retrieved photo in \cref{fig: hs-fairness}. Humans can identify the misclassified category $62.4\%(66.3\%)$ and the incorrectly retrieved photo $39.1\%(37.2\%)$ for SLA (P-SLA).

\begin{figure}[h]
    \centering
    \includegraphics[width=\linewidth,trim={0 2mm 0 6mm}]{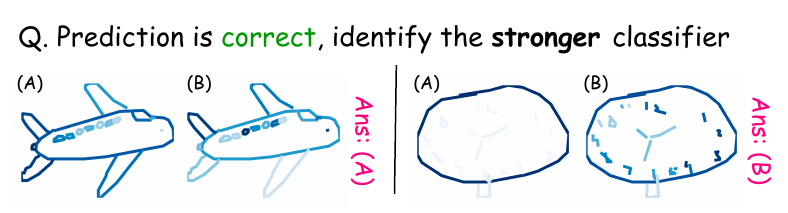}
    \caption{Evaluating trustworthiness: We present the stroke attributions (SLA/P-SLA) from two sketch classifiers (strong and weak), and ask users to identify the strong/weak model.}
    \label{fig: hs-trustworthiness}
    \vspace{-6mm}
\end{figure}

\keypoint{Evaluting Trustworthiness:} Determining trust in individual predictions is important when used for decision-making (e.g., medical diagnosis \cite{kulesza2015debugging}). We train two copies of the same sketch classifier \cite{yu2015sketch}, a strong classifier with $73.8\%$ accuracy on TU-Berlin \cite{berlin} and a weak one reaching $57.1\%$. We present the stroke attribution (SLA/P-SLA) for both models and ask users to identify the strong/weak classifier, as shown in \cref{fig: hs-trustworthiness}. Humans identify the stronger classifier $71.4\%(68.7\%)$ of the time for SLA (P-SLA).

\section{Conclusion}
This work emphasises the pivotal role of strokes in human-drawn sketches, offering unique insights compared to pixel-based images. Our lightweight explainability solution seamlessly integrates with pre-trained models, addressing rasterisation challenges and contributing to diverse sketch-related tasks. Through applications in Retrieval, Generation, Assisted Drawing, and Sketch Adversarial Attack, our model showcases adaptability and significance. The proposed stroke-level attribution provides nuanced insights into model behaviour, underscoring the importance of explainability in bridging human expression with model predictions in the evolving field of sketch interpretation.

{
    \small
    \bibliographystyle{ieeenat_fullname}
    \bibliography{main}

\begin{thebibliography}{124}
\providecommand{\natexlab}[1]{#1}
\providecommand{\url}[1]{\texttt{#1}}
\expandafter\ifx\csname urlstyle\endcsname\relax
  \providecommand{\doi}[1]{doi: #1}\else
  \providecommand{\doi}{doi: \begingroup \urlstyle{rm}\Url}\fi

\bibitem[Adebayo et~al.(2018)Adebayo, Gilmer, Muelly, Goodfellow, Hardt, and Kim]{adebayo2018sanity}
Julius Adebayo, Justin Gilmer, Michael Muelly, Ian Goodfellow, Moritz Hardt, and Been Kim.
\newblock Sanity checks for saliency maps.
\newblock In \emph{NeurIPS}, 2018.

\bibitem[Adebayo et~al.(2020)Adebayo, Muelly, Liccardi, and Kim]{adebayo2020debugging}
Julius Adebayo, Michael Muelly, Ilaria Liccardi, and Been Kim.
\newblock Debugging tests for model explanations.
\newblock \emph{arXiv preprint arXiv:2011.05429}, 2020.

\bibitem[Akhtar et~al.(2021)Akhtar, Mian, Kardan, and Shah]{akhtar2021attack}
Naveed Akhtar, Ajmal Mian, Navid Kardan, and Mubarak Shah.
\newblock Advances in adversarial attacks and defenses in computer vision: A survey.
\newblock \emph{arXiv preprint arXiv:2108.00401}, 2021.

\bibitem[Almeda et~al.(2023)Almeda, Zamfirescu-Pereira, Kim, Rathnam, and Hartmann]{dreamsheets}
Shm~Garanganao Almeda, J.D. Zamfirescu-Pereira, Kyu~Won Kim, Pradeep~Mani Rathnam, and Bjoern Hartmann.
\newblock Prompting for discovery: Flexible sense-making for ai art-making with dreamsheets.
\newblock \emph{arXiv preprint arXiv:2310.09985}, 2023.

\bibitem[Ancona et~al.(2017{\natexlab{a}})Ancona, Ceolini, {\"O}ztireli, and Gross]{ancona2017towards}
Marco Ancona, Enea Ceolini, Cengiz {\"O}ztireli, and Markus Gross.
\newblock Towards better understanding of gradient-based attribution methods for deep neural networks.
\newblock \emph{arXiv preprint arXiv:1711.06104}, 2017{\natexlab{a}}.

\bibitem[Ancona et~al.(2017{\natexlab{b}})Ancona, Ceolini, {\"O}ztireli, and Gross]{ancona2017unified}
Marco Ancona, Enea Ceolini, Cengiz {\"O}ztireli, and Markus Gross.
\newblock A unified view of gradient-based attribution methods for deep neural networks.
\newblock In \emph{NIPSW}, 2017{\natexlab{b}}.

\bibitem[Arnab et~al.(2018)Arnab, Miksik, and Torr]{arnab2018attack}
Anurag Arnab, Ondrej Miksik, and Philip~H.S. Torr.
\newblock On the robustness of semantic segmentation models to adversarial attacks.
\newblock In \emph{CVPR}, 2018.

\bibitem[Bach et~al.(2015)Bach, Binder, Montavon, Klauschen, M{\"u}ller, and Samek]{bach2015lrp}
Sebastian Bach, Alexander Binder, Gr{\'e}goire Montavon, Frederick Klauschen, Klaus-Robert M{\"u}ller, and Wojciech Samek.
\newblock On pixel-wise explanations for non-linear classifier decisions by layer-wise relevance propagation.
\newblock \emph{PloS one}, 2015.

\bibitem[Bandyopadhyay et~al.(2024{\natexlab{a}})Bandyopadhyay, Bhunia, Chowdhury, Sain, Xiang, Hospedales, and Song]{bandyopadhyay2024INR}
Hmrishav Bandyopadhyay, Ayan~Kumar Bhunia, Pinaki~Nath Chowdhury, Aneeshan Sain, Tao Xiang, Timothy Hospedales, and Yi-Zhe Song.
\newblock Sketchinr: A first look into sketches as implicit neural representations.
\newblock In \emph{CVPR}, 2024{\natexlab{a}}.

\bibitem[Bandyopadhyay et~al.(2024{\natexlab{b}})Bandyopadhyay, Koley, Das, Bhunia, Sain, Chowdhury, Xiang, and Song]{bandyopadhyay2024doodle}
Hmrishav Bandyopadhyay, Subhadeep Koley, Ayan Das, Ayan~Kumar Bhunia, Aneeshan Sain, Pinaki~Nath Chowdhury, Tao Xiang, and Yi-Zhe Song.
\newblock Doodle your 3d: From abstract freehand sketches to precise 3d shapes.
\newblock In \emph{CVPR}, 2024{\natexlab{b}}.

\bibitem[Bhunia et~al.(2020{\natexlab{a}})Bhunia, Das, Muhammad, Yang, Hospedales, Xiang, Gryaditskaya, and Song]{bhunia2020pixelor}
Ayan~Kumar Bhunia, Ayan Das, Umar~Riaz Muhammad, Yongxin Yang, Timothy~M Hospedales, Tao Xiang, Yulia Gryaditskaya, and Yi-Zhe Song.
\newblock Pixelor: A competitive sketching ai agent. so you think you can sketch?
\newblock \emph{ACM TOG}, 2020{\natexlab{a}}.

\bibitem[Bhunia et~al.(2020{\natexlab{b}})Bhunia, Yang, Hospedales, Xiang, and Song]{bhunia2020sketch}
Ayan~Kumar Bhunia, Yongxin Yang, Timothy~M Hospedales, Tao Xiang, and Yi-Zhe Song.
\newblock Sketch less for more: On-the-fly fine-grained sketch based image retrieval.
\newblock In \emph{CVPR}, 2020{\natexlab{b}}.

\bibitem[Bhunia et~al.(2021)Bhunia, Chowdhury, Sain, Yang, Xiang, and Song]{bhunia2021more}
Ayan~Kumar Bhunia, Pinaki~Nath Chowdhury, Aneeshan Sain, Yongxin Yang, Tao Xiang, and Yi-Zhe Song.
\newblock More photos are all you need: Semi-supervised learning for fine-grained sketch based image retrieval.
\newblock In \emph{CVPR}, 2021.

\bibitem[Bhunia et~al.(2022{\natexlab{a}})Bhunia, Gajjala, Koley, Kundu, Sain, Xiang, and Song]{bhunia2022doodle}
Ayan~Kumar Bhunia, Viswanatha~Reddy Gajjala, Subhadeep Koley, Rohit Kundu, Aneeshan Sain, Tao Xiang, and Yi-Zhe Song.
\newblock Doodle it yourself: Class incremental learning by drawing a few sketches.
\newblock In \emph{CVPR}, 2022{\natexlab{a}}.

\bibitem[Bhunia et~al.(2022{\natexlab{b}})Bhunia, Koley, Khilji, Sain, Chowdhury, Xiang, and Song]{strokesubset}
Ayan~Kumar Bhunia, Subhadeep Koley, Abdullah Faiz Ur~Rahman Khilji, Aneeshan Sain, Pinaki~Nath Chowdhury, Tao Xiang, and Yi-Zhe Song.
\newblock Sketching without worrying: Noise-tolerant sketch-based image retrieval.
\newblock In \emph{CVPR}, 2022{\natexlab{b}}.

\bibitem[B\:ohle et~al.(2022)B\:ohle, Fritz, and Schiele]{B-cos-networks}
Moritz B\:ohle, Mario Fritz, and Bernt Schiele.
\newblock B-cos: Networks: Alignment is all we need for interpretability.
\newblock In \emph{CVPR}, 2022.

\bibitem[Brendel and Bethge(2019)]{brendel2019approximating}
Wieland Brendel and Matthias Bethge.
\newblock Approximating cnns with bag-of-local-features models works surprisingly well on imagenet.
\newblock \emph{arXiv preprint arXiv:1904.00760}, 2019.

\bibitem[Cao et~al.(2015)Cao, Liu, Yang, Yu, Wang, Wang, Huang, Wang, Huang, Xu, et~al.]{cao2015look}
Chunshui Cao, Xianming Liu, Yi Yang, Yinan Yu, Jiang Wang, Zilei Wang, Yongzhen Huang, Liang Wang, Chang Huang, Wei Xu, et~al.
\newblock Look and think twice: Capturing top-down visual attention with feedback convolutional neural networks.
\newblock In \emph{ICCV}, 2015.

\bibitem[Chen et~al.(2019)Chen, Li, Tao, Barnett, Rudin, and Su]{chen2019looks}
Chaofan Chen, Oscar Li, Daniel Tao, Alina Barnett, Cynthia Rudin, and Jonathan~K Su.
\newblock This looks like that: deep learning for interpretable image recognition.
\newblock In \emph{NeurIPS}, 2019.

\bibitem[Chen et~al.(2020{\natexlab{a}})Chen, Su, Gao, Xia, and Fu]{chen2020deepfacedrawing}
Shu-Yu Chen, Wanchao Su, Lin Gao, Shihong Xia, and Hongbo Fu.
\newblock Deepfacedrawing: Deep generation of face images from sketches.
\newblock \emph{ACM TOG}, 2020{\natexlab{a}}.

\bibitem[Chen and Hays(2018)]{chen2018sketchygan}
Wengling Chen and James Hays.
\newblock Sketchygan: Towards diverse and realistic sketch to image synthesis.
\newblock In \emph{ICCV}, 2018.

\bibitem[Chen et~al.(2020{\natexlab{b}})Chen, Yan, Zheng, Jiang, Xia, Zhao, and Ji]{chen2020attack}
Xuesong Chen, Xiyu Yan, Feng Zheng, Yong Jiang, Shu-Tao Xia, Yong Zhao, and Rongrong Ji.
\newblock One-shot adversarial attacks on visual tracking with dual attention.
\newblock In \emph{CVPR}, 2020{\natexlab{b}}.

\bibitem[Chowdhury et~al.(2022{\natexlab{a}})Chowdhury, Bhunia, Gajjala, Sain, Xiang, and Song]{chowdhury2022partially}
Pinaki~Nath Chowdhury, Ayan~Kumar Bhunia, Viswanatha~Reddy Gajjala, Aneeshan Sain, Tao Xiang, and Yi-Zhe Song.
\newblock Partially does it: Towards scene-level fg-sbir with partial input.
\newblock In \emph{CVPR}, 2022{\natexlab{a}}.

\bibitem[Chowdhury et~al.(2022{\natexlab{b}})Chowdhury, Sain, Gryaditskaya, Bhunia, Xiang, and Song]{fscoco}
Pinaki~Nath Chowdhury, Aneeshan Sain, Yulia Gryaditskaya, Ayan~Kumar Bhunia, Tao Xiang, and Yi-Zhe Song.
\newblock Fs-coco: Towards understanding of freehand sketches of common objects in context.
\newblock In \emph{ECCV}, 2022{\natexlab{b}}.

\bibitem[Chowdhury et~al.(2022{\natexlab{c}})Chowdhury, Wang, Ceylan, Song, and Gryaditskaya]{chowdhury20223Dsynthesis}
Pinaki~Nath Chowdhury, Tuanfeng Wang, Duygu Ceylan, Yi-Zhe Song, and Yulia Gryaditskaya.
\newblock Garment ideation: Iterative view-aware sketch-based garment modeling.
\newblock In \emph{3DV}, 2022{\natexlab{c}}.

\bibitem[Collomosse et~al.(2019)Collomosse, Bui, and Jin]{collomosse2019livesketch}
John Collomosse, Tu Bui, and Hailin Jin.
\newblock Livesketch: Query perturbations for guided sketch-based visual search.
\newblock In \emph{CVPR}, 2019.

\bibitem[Dabkowski and Gal(2017)]{dabkowski2017real}
Piotr Dabkowski and Yarin Gal.
\newblock Real time image saliency for black box classifiers.
\newblock In \emph{NeurIPS}, 2017.

\bibitem[Das et~al.(2020)Das, Yang, Hospedales, Xiang, and Song]{das2020beziersketch}
Ayan Das, Yongxin Yang, Timothy Hospedales, Tao Xiang, and Yi-Zhe Song.
\newblock B{\'e}ziersketch: A generative model for scalable vector sketches.
\newblock In \emph{ECCV}, 2020.

\bibitem[Dey et~al.(2019)Dey, Riba, Dutta, Llados, and Song]{dey2019doodle}
Sounak Dey, Pau Riba, Anjan Dutta, Josep Llados, and Yi-Zhe Song.
\newblock Doodle to search: Practical zero-shot sketch-based image retrieval.
\newblock In \emph{CVPR}, 2019.

\bibitem[Dhamdhere et~al.(2018)Dhamdhere, Sundararajan, and Yan]{dhamdhere2018important}
Kedar Dhamdhere, Mukund Sundararajan, and Qiqi Yan.
\newblock How important is a neuron?
\newblock \emph{arXiv preprint arXiv:1805.12233}, 2018.

\bibitem[Dominici et~al.(2023)Dominici, Barbiero, Magister, Pietro, and Simidjievski]{dominici2023sharcs}
Gabriele Dominici, Pietro Barbiero, Lucie~Charlotte Magister, Li\`o Pietro, and Nikola Simidjievski.
\newblock Sharcs: Shared concept space for explainable multimodal learning.
\newblock \emph{arXiv preprint arXiv:2307.00316}, 2023.

\bibitem[Dutta and Akata(2019)]{dutta2019semantically}
Anjan Dutta and Zeynep Akata.
\newblock Semantically tied paired cycle consistency for zero-shot sketch-based image retrieval.
\newblock In \emph{CVPR}, 2019.

\bibitem[Eitz et~al.(2012)Eitz, Hays, and Alexa]{berlin}
Mathias Eitz, James Hays, and Marc Alexa.
\newblock How do humans sketch objects?
\newblock \emph{ACM TOG}, 2012.

\bibitem[Erhan et~al.(2009)Erhan, Bengio, Courville, and Vincent]{erhan2009visualizing}
Dumitru Erhan, Yoshua Bengio, Aaron Courville, and Pascal Vincent.
\newblock Visualizing higher-layer features of a deep network.
\newblock \emph{Technical Report, Univeristé de Montréal}, 2009.

\bibitem[Fel et~al.(2021)Fel, Colin, Cad{\`e}ne, and Serre]{fel2021cannot}
Thomas Fel, Julien Colin, R{\'e}mi Cad{\`e}ne, and Thomas Serre.
\newblock What i cannot predict, i do not understand: A human-centered evaluation framework for explainability methods.
\newblock \emph{arXiv preprint arXiv:2112.04417}, 2021.

\bibitem[Fong et~al.(2019)Fong, Patrick, and Vedaldi]{fong2019understanding}
Ruth Fong, Mandela Patrick, and Andrea Vedaldi.
\newblock Understanding deep networks via extremal perturbations and smooth masks.
\newblock In \emph{ICCV}, 2019.

\bibitem[Fong and Vedaldi(2017)]{fong2017interpretable}
Ruth~C Fong and Andrea Vedaldi.
\newblock Interpretable explanations of black boxes by meaningful perturbation.
\newblock In \emph{ICCV}, 2017.

\bibitem[Gao et~al.(2020)Gao, Liu, Xu, Wang, Liu, and Zou]{gao2020sketchycoco}
Chengying Gao, Qi Liu, Qi Xu, Limin Wang, Jianzhuang Liu, and Changqing Zou.
\newblock Sketchycoco: Image generation from freehand scene sketches.
\newblock In \emph{CVPR}, 2020.

\bibitem[Guillard et~al.(2021)Guillard, Remelli, Yvernay, and Fua]{guillard2021sketch2mesh}
Benoit Guillard, Edoardo Remelli, Pierre Yvernay, and Pascal Fua.
\newblock Sketch2mesh: Reconstructing and editing 3d shapes from sketches.
\newblock In \emph{CVPR}, 2021.

\bibitem[Ha and Eck(2018)]{ha2017neural}
David Ha and Douglas Eck.
\newblock A neural representation of sketch drawings.
\newblock In \emph{ICLR}, 2018.

\bibitem[He et~al.(2016)He, Zhang, Ren, and Sun]{HeResNet}
Kaiming He, Xiangyu Zhang, Shaoqing Ren, and Jian Sun.
\newblock Deep residual learning for image recognition.
\newblock In \emph{CVPR}, 2016.

\bibitem[He et~al.(2020)He, Rahimian, Schiele, and Fritz]{he2020attack}
Yang He, Shadi Rahimian, Bernt Schiele, and Mario Fritz.
\newblock Segmentations-leak: Membership inference attacks and defenses in semantic image segmentation.
\newblock In \emph{ECCV}, 2020.

\bibitem[Hertzmann(2020)]{hertzmann2020line}
Aaron Hertzmann.
\newblock Why do line drawings work? a realism hypothesis.
\newblock \emph{Perception}, 2020.

\bibitem[Holinaty et~al.(2021)Holinaty, Jacobson, and Chevalier]{holinaty2021drawing}
Josh Holinaty, Alec Jacobson, and Fanny Chevalier.
\newblock Supportingreferenceimageryfordigitaldrawing.
\newblock In \emph{ICCVW}, 2021.

\bibitem[Hu et~al.(2020)Hu, Li, Yang, Hospedales, and Song]{hu2020sketch}
Conghui Hu, Da Li, Yongxin Yang, Timothy~M Hospedales, and Yi-Zhe Song.
\newblock Sketch-a-segmenter: Sketch-based photo segmenter generation.
\newblock \emph{IEEE TIP}, 2020.

\bibitem[Huynh-Thu et~al.(2010)Huynh-Thu, Garcia, Speranza, Corriveau, and Raake]{thu2010mos}
Quan Huynh-Thu, Marie-Neige Garcia, Filippo Speranza, Philip Corriveau, and Alexander Raake.
\newblock Study of ratling scales for subjective quality assessment of high definition video.
\newblock \emph{IEEE TBC}, 2010.

\bibitem[Jang et~al.(2017)Jang, Gu, and Poole]{jang2016categorical}
Eric Jang, Shixiang Gu, and Ben Poole.
\newblock Categorical reparameterization with gumbel-softmax.
\newblock In \emph{ICLR}, 2017.

\bibitem[Jia et~al.(2020)Jia, Lu, Shen, Chen, Chen, Zhong, and Wei]{jia2020attack}
Yunhan Jia, Yantao Lu, Junjie Shen, Qi~Alfred Chen, Hao Chen, Zhenyu Zhong, and Tao Wei.
\newblock Fooling detection alone is not enough: Adversarial attack against multiple object tracking.
\newblock In \emph{ICLR}, 2020.

\bibitem[Jia et~al.(2023)Jia, Ben-Michael, and Imai]{jia2023bayesian}
Zeyang Jia, Eli Ben-Michael, and Kosuke Imai.
\newblock Bayesian safe policy learning with chance constrained optimization: Application to military security assessment during the vietnam war.
\newblock \emph{arXiv preprint arXiv:2307.08840}, 2023.

\bibitem[Kang et~al.(2023)Kang, Zhu, Zhang, Park, Shechtman, Paris, and Park]{GigaGan}
Minguk Kang, Jun-Yan Zhu, Richard Zhang, Jaesik Park, Eli Shechtman, Sylvain Paris, and Taesung Park.
\newblock Scaling up gans for text-to-image synthesis.
\newblock In \emph{CVPR}, 2023.

\bibitem[Kapishnikov et~al.(2019)Kapishnikov, Bolukbasi, Vi{\'e}gas, and Terry]{kapishnikov2019xrai}
Andrei Kapishnikov, Tolga Bolukbasi, Fernanda Vi{\'e}gas, and Michael Terry.
\newblock Xrai: Better attributions through regions.
\newblock In \emph{ICCV}, 2019.

\bibitem[Kapishnikov et~al.(2021)Kapishnikov, Venugopalan, Avci, Wedin, Terry, and Bolukbasi]{kapishnikov2021guided}
Andrei Kapishnikov, Subhashini Venugopalan, Besim Avci, Ben Wedin, Michael Terry, and Tolga Bolukbasi.
\newblock Guided integrated gradients: An adaptive path method for removing noise.
\newblock In \emph{CVPR}, 2021.

\bibitem[Karras et~al.(2019)Karras, Laine, and Aila]{style-gan}
Tero Karras, Samuli Laine, and Timo Aila.
\newblock A style-based generator architecture for generative adversarial networks.
\newblock In \emph{CVPR}, 2019.

\bibitem[Kim et~al.(2022)Kim, Meister, Ramaswamy, Fong, and Russakovsky]{kim2022hive}
Sunnie S.~Y. Kim, Nicole Meister, Vikram~V. Ramaswamy, Ruth Fong, and Olga Russakovsky.
\newblock Hive: Evaluating the human interpretability of visual explanations.
\newblock In \emph{ECCV}, 2022.

\bibitem[Kindermans et~al.(2019)Kindermans, Hooker, Adebayo, Alber, Sch{\"u}tt, D{\"a}hne, Erhan, and Kim]{kindermans2019reliability}
Pieter-Jan Kindermans, Sara Hooker, Julius Adebayo, Maximilian Alber, Kristof~T Sch{\"u}tt, Sven D{\"a}hne, Dumitru Erhan, and Been Kim.
\newblock The (un) reliability of saliency methods.
\newblock In \emph{Explainable AI: Interpreting, Explaining and Visualizing Deep Learning}, 2019.

\bibitem[Koley et~al.(2023)Koley, Bhunia, Sain, Chowdhury, Xiang, and Song]{koley2023picture}
Subhadeep Koley, Ayan~Kumar Bhunia, Aneeshan Sain, Pinaki~Nath Chowdhury, Tao Xiang, and Yi-Zhe Song.
\newblock Picture that sketch: Photorealistic image generation from abstract sketches.
\newblock In \emph{CVPR}, 2023.

\bibitem[Kulesza et~al.(2015{\natexlab{a}})Kulesza, Burnett, Wong, and Stumpf]{kulesza2015bayesian}
Todd Kulesza, Margaret Burnett, Weng-Keen Wong, and Simone Stumpf.
\newblock Principles of explanatory debugging to personalize interactive machine learning.
\newblock In \emph{IUI}, 2015{\natexlab{a}}.

\bibitem[Kulesza et~al.(2015{\natexlab{b}})Kulesza, Burnett, Wong, and Stumpf]{kulesza2015debugging}
Todd Kulesza, Margaret Burnett, Weng-Keen Wong, and Simone Stumpf.
\newblock Principles of explanatory debugging to personalize interactive machine learning.
\newblock In \emph{IUI}, 2015{\natexlab{b}}.

\bibitem[Lage et~al.(2019)Lage, Chen, He, Narayanan, Kim, Gershman, and Doshi-Velez]{lage2019evaluation}
Isaac Lage, Emily Chen, Jeffrey He, Menaka Narayanan, Been Kim, Sam Gershman, and Finale Doshi-Velez.
\newblock An evaluation of the human-interpretability of explanation.
\newblock \emph{arXiv preprint arXiv:1902.00006}, 2019.

\bibitem[Li et~al.(2019)Li, Lin, Mech, Yumer, and Ramanan]{li2019photo}
Mengtian Li, Zhe Lin, Radomir Mech, Ersin Yumer, and Deva Ramanan.
\newblock Photo-sketching: Inferring contour drawings from images.
\newblock In \emph{WACV}, 2019.

\bibitem[Li et~al.(2020)Li, Deng, Li, Yan, Gao, and Huang]{li2020tranferable}
Maosen Li, Cheng Deng, Tengjiao Li, Junchi Yan, Xinbo Gao, and Heng Huang.
\newblock Towards transferable targeted attack.
\newblock In \emph{CVPR}, 2020.

\bibitem[Lin et~al.(2023)Lin, Li, Li, Hospedales, Song, and Qi]{yongangXAI}
Fengyin Lin, Mingkang Li, Da Li, Timothy Hospedales, Yi-Zhe Song, and Yonggang Qi.
\newblock Zero-shot everything sketch-based image retrieval, and in explainable style.
\newblock In \emph{CVPR}, 2023.

\bibitem[Liu et~al.(2021)Liu, Wan, Huang, Song, Han, Liao, Jiang, and Liu]{liu2021deflocnet}
Hongyu Liu, Ziyu Wan, Wei Huang, Yibing Song, Xintong Han, Jing Liao, Bin Jiang, and Wei Liu.
\newblock Deflocnet: Deep image editing via flexible low-level controls.
\newblock In \emph{CVPR}, 2021.

\bibitem[Liu et~al.(2017)Liu, Shen, Shen, Liu, and Shao]{liu2017deep}
Li Liu, Fumin Shen, Yuming Shen, Xianglong Liu, and Ling Shao.
\newblock Deep sketch hashing: Fast free-hand sketch-based image retrieval.
\newblock In \emph{CVPR}, 2017.

\bibitem[Mou et~al.(2023)Mou, Wang, Xie, Wu, Zhang, Qi, Shan, and Qie]{t2i-adapter}
Chong Mou, Xintao Wang, Liangbin Xie, Yanze Wu, Jian Zhang, Zhongang Qi, Ying Shan, and Xiaohu Qie.
\newblock T2i-adapter: Learning adapters to dig out more controllable ability for text-to-image diffusion models.
\newblock \emph{arXiv preprint arXiv:2302.08453}, 2023.

\bibitem[Nauta et~al.(2021)Nauta, Van~Bree, and Seifert]{nauta2021neural}
Meike Nauta, Ron Van~Bree, and Christin Seifert.
\newblock Neural prototype trees for interpretable fine-grained image recognition.
\newblock In \emph{CVPR}, 2021.

\bibitem[Nguyen et~al.(2016)Nguyen, Yosinski, and Clune]{nguyen2016multifaceted}
Anh Nguyen, Jason Yosinski, and Jeff Clune.
\newblock Multifaceted feature visualization: Uncovering the different types of features learned by each neuron in deep neural networks.
\newblock \emph{ICMLW}, 2016.

\bibitem[Olah et~al.(2017)Olah, Mordvintsev, and Schubert]{olah2017feature}
Chris Olah, Alexander Mordvintsev, and Ludwig Schubert.
\newblock Feature visualization.
\newblock \emph{Distill}, 2017.

\bibitem[Pang et~al.(2019)Pang, Li, Yang, Zhang, Hospedales, Xiang, and Song]{pang2019generalising}
Kaiyue Pang, Ke Li, Yongxin Yang, Honggang Zhang, Timothy~M Hospedales, Tao Xiang, and Yi-Zhe Song.
\newblock Generalising fine-grained sketch-based image retrieval.
\newblock In \emph{CVPR}, 2019.

\bibitem[Petsiuk et~al.(2018)Petsiuk, Das, and Saenko]{petsiuk2018rise}
Vitali Petsiuk, Abir Das, and Kate Saenko.
\newblock Rise: Randomized input sampling for explanation of black-box models.
\newblock \emph{arXiv preprint arXiv:1806.07421}, 2018.

\bibitem[Phan et~al.(2021)Phan, Mannan, and Heide]{phan2021adversarial}
Buu Phan, Fahim Mannan, and Felix Heide.
\newblock Adversarial imaging pipelines.
\newblock In \emph{CVPR}, 2021.

\bibitem[Qu et~al.(2023)Qu, Gryaditskaya1, Li, Pang, Xiang, and Song]{sketchXAI}
Zhiyu Qu, Yulia Gryaditskaya1, Ke Li, Kaiyue Pang, Tao Xiang, and Yi-Zhe Song.
\newblock Sketchxai: A first look at explainability for human sketches.
\newblock In \emph{CVPR}, 2023.

\bibitem[Radford et~al.(2021)Radford, Kim, Hallacy, Ramesh, Goh, Agarwal, Sastry, Askell, Mishkin, Clark, Krueger, and Sutskever]{CLIP}
Alec Radford, Jong~Wook Kim, Chris Hallacy, Aditya Ramesh, Gabriel Goh, Sandhini Agarwal, Girish Sastry, Amanda Askell, Pamela Mishkin, Jack Clark, Gretchen Krueger, and Ilya Sutskever.
\newblock Learning transferable visual models from natural language supervision.
\newblock In \emph{ICML}, 2021.

\bibitem[Rao et~al.(2022)Rao, B{\"o}hle, and Schiele]{rao2022towards}
Sukrut Rao, Moritz B{\"o}hle, and Bernt Schiele.
\newblock Towards better understanding attribution methods.
\newblock In \emph{CVPR}, 2022.

\bibitem[Ribeiro et~al.(2016)Ribeiro, Singh, and Guestrin]{LIME}
Marco~T. Ribeiro, Sameer Singh, and Carlos Guestrin.
\newblock ``why should i trust you?'' explaining the predictions of any classifier.
\newblock In \emph{KDD}, 2016.

\bibitem[Rombach et~al.(2022)Rombach, Blattmann, Lorenz, Esser, and Ommer]{stable-diffusion}
Robin Rombach, Andreas Blattmann, Dominik Lorenz, Patrick Esser, and Bj\:orn Ommer.
\newblock High-resolution image synthesis with latent diffusion.
\newblock In \emph{CVPR}, 2022.

\bibitem[Rudin(2019)]{rudin2019blackbox}
Cynthia Rudin.
\newblock Stop explaining black box machine learning models for high stakes decisions and use interpretable models instead.
\newblock \emph{Nature Machine Intelligence}, 2019.

\bibitem[Sain et~al.(2021)Sain, Bhunia, Yang, Xiang, and Song]{sain2021stylemeup}
Aneeshan Sain, Ayan~Kumar Bhunia, Yongxin Yang, Tao Xiang, and Yi-Zhe Song.
\newblock Stylemeup: Towards style-agnostic sketch-based image retrieval.
\newblock In \emph{CVPR}, 2021.

\bibitem[Sain et~al.(2023)Sain, Bhunia, Chowdhury, Koley, Xiang, and Song]{sain2023clip}
Aneeshan Sain, Ayan~Kumar Bhunia, Pinaki~Nath Chowdhury, Subhadeep Koley, Tao Xiang, and Yi-Zhe Song.
\newblock Clip for all things zero-shot sketch-based image retrieval, fine-grained or not.
\newblock In \emph{CVPR}, 2023.

\bibitem[Samek et~al.(2016)Samek, Binder, Montavon, Lapuschkin, and M{\"u}ller]{samek2016evaluating}
Wojciech Samek, Alexander Binder, Gr{\'e}goire Montavon, Sebastian Lapuschkin, and Klaus-Robert M{\"u}ller.
\newblock Evaluating the visualization of what a deep neural network has learned.
\newblock \emph{IEEE TNNLS}, 2016.

\bibitem[Sangkloy et~al.(2016)Sangkloy, Burnell, Ham, and Hays]{sangkloy2016sketchy}
Patsorn Sangkloy, Nathan Burnell, Cusuh Ham, and James Hays.
\newblock The sketchy database: learning to retrieve badly drawn bunnies.
\newblock \emph{ACM TOG}, 2016.

\bibitem[Schulman et~al.(2017)Schulman, Wolski, Dhariwal, Radford, and Klimov]{PPO}
John Schulman, Filip Wolski, Prafulla Dhariwal, Alec Radford, and Oleg Klimov.
\newblock Pronimal policy optimization algorithms.
\newblock \emph{arXiv preprint arXiv:1707.06347}, 2017.

\bibitem[Selvaraju et~al.(2017{\natexlab{a}})Selvaraju, Cogswell, Das, Vedantam, Parikh, and Batra]{gradCAM}
Ramprasaath~R. Selvaraju, Michael Cogswell, Abhishek Das, Ramakrishna Vedantam, Devi Parikh, and Dhruv Batra.
\newblock Grad-cam: Visual explanations from deep networks via gradient-based localisation.
\newblock In \emph{ICCV}, 2017{\natexlab{a}}.

\bibitem[Selvaraju et~al.(2017{\natexlab{b}})Selvaraju, Cogswell, Das, Vedantam, Parikh, and Batra]{selvaraju2017grad}
Ramprasaath~R Selvaraju, Michael Cogswell, Abhishek Das, Ramakrishna Vedantam, Devi Parikh, and Dhruv Batra.
\newblock Grad-cam: Visual explanations from deep networks via gradient-based localization.
\newblock In \emph{ICCV}, 2017{\natexlab{b}}.

\bibitem[Shah et~al.(2021)Shah, Jain, and Netrapalli]{shah2021input}
Harshay Shah, Prateek Jain, and Praneeth Netrapalli.
\newblock Do input gradients highlight discriminative features?
\newblock In \emph{NeurIPS}, 2021.

\bibitem[Sheikh and Bovik(2006)]{sheikh2006quality}
Hamid~Rahim Sheikh and Alan~C. Bovik.
\newblock Image information and visual quality.
\newblock \emph{IEEE TIP}, 2006.

\bibitem[Shen et~al.(2018)Shen, Liu, Shen, and Shao]{shen2018zero}
Yuming Shen, Li Liu, Fumin Shen, and Ling Shao.
\newblock Zero-shot sketch-image hashing.
\newblock In \emph{ICCV}, 2018.

\bibitem[Shrikumar et~al.(2017)Shrikumar, Greenside, and Kundaje]{shrikumar2017deeplift}
Avanti Shrikumar, Peyton Greenside, and Anshul Kundaje.
\newblock Learning important features through propagating activation differences.
\newblock In \emph{ICML}, 2017.

\bibitem[Simonyan et~al.(2014)Simonyan, Vedaldi, and Zisserman]{simonyan2013}
Karen Simonyan, Andrea Vedaldi, and Andrew Zisserman.
\newblock Deep inside convolutional networks: Visualising image classification models and saliency maps.
\newblock In \emph{ICLRW}, 2014.

\bibitem[Sixt et~al.(2022)Sixt, Schuessler, Popescu, Wei{\ss}, and Landgraf]{sixt2022users}
Leon Sixt, Martin Schuessler, Oana-Iuliana Popescu, Philipp Wei{\ss}, and Tim Landgraf.
\newblock Do users benefit from interpretable vision? a user study, baseline, and dataset.
\newblock \emph{arXiv preprint arXiv:2204.11642}, 2022.

\bibitem[Smilkov et~al.(2017)Smilkov, Thorat, Kim, Vi{\'e}gas, and Wattenberg]{smilkov2017smoothgrad}
Daniel Smilkov, Nikhil Thorat, Been Kim, Fernanda Vi{\'e}gas, and Martin Wattenberg.
\newblock Smoothgrad: removing noise by adding noise.
\newblock \emph{arXiv preprint arXiv:1706.03825}, 2017.

\bibitem[Springenberg et~al.(2014)Springenberg, Dosovitskiy, Brox, and Riedmiller]{springenberg2014guided_backprop}
Jost~Tobias Springenberg, Alexey Dosovitskiy, Thomas Brox, and Martin Riedmiller.
\newblock Striving for simplicity: The all convolutional net.
\newblock \emph{arXiv preprint arXiv:1412.6806}, 2014.

\bibitem[Springenberg et~al.(2015)Springenberg, Dosovitskiy, Brox, and Riedmiller]{tobias2015simplicity}
Jost~Tobias Springenberg, Alexey Dosovitskiy, Thomas Brox, and Martin Riedmiller.
\newblock Striving for simplicity: The all convolutional net.
\newblock In \emph{ICLRW}, 2015.

\bibitem[Sundararajan et~al.(2017)Sundararajan, Taly, and Yan]{sundararajan2017axiomatic}
Mukund Sundararajan, Ankur Taly, and Qiqi Yan.
\newblock Axiomatic attribution for deep networks.
\newblock In \emph{ICML}, 2017.

\bibitem[Szegedy et~al.(2013{\natexlab{a}})Szegedy, Zaremba, Sutskever, Bruna, Erhan, Goodfellow, and Fergus]{szegedy2013}
Christian Szegedy, Wojciech Zaremba, Ilyua Sutskever, Joan Bruna, Dumitru Erhan, Ian Goodfellow, and Rob Fergus.
\newblock Intriguing properties of neural networks.
\newblock \emph{arXiv preprint arXiv:1312.6199}, 2013{\natexlab{a}}.

\bibitem[Szegedy et~al.(2013{\natexlab{b}})Szegedy, Zaremba, Sutskever, Bruna, Erhan, Goodfellow, and Fergus]{szegedy2013attacks}
Christian Szegedy, Wojciech Zaremba, Ilya Sutskever, Joan Bruna, Dumitru Erhan, Ian Goodfellow, and Rob Fergus.
\newblock Intriguing properties of neural networks.
\newblock \emph{arXiv preprint arXiv:1312.6199}, 2013{\natexlab{b}}.

\bibitem[Tashiro et~al.(2020)Tashiro, Song, and Ermon]{tashiro2020diversity}
Yusuke Tashiro, Yang Song, and Stefano Ermon.
\newblock Diversity can be transferred: Output diversification for white- and black-box attacks.
\newblock In \emph{NeurIPS}, 2020.

\bibitem[Tripathi et~al.(2020)Tripathi, Dani, Mishra, and Chakraborty]{tripathi2020sketch}
Aditay Tripathi, Rajath~R Dani, Anand Mishra, and Anirban Chakraborty.
\newblock Sketch-guided object localization in natural images.
\newblock In \emph{ECCV}, 2020.

\bibitem[Tu et~al.(2020)Tu, Ren, Manivasagam, Liang, Yang, Du, Cheng, and Urtasun]{tu2020attack}
James Tu, Mengye Ren, Siva Manivasagam, Ming Liang, Bin Yang, Richard Du, Frank Cheng, and Raquel Urtasun.
\newblock Physically realizable adversarial examples for lidar object detection.
\newblock In \emph{CVPR}, 2020.

\bibitem[Turb\'e et~al.(2023)Turb\'e, Bjelogrlic, Lovis, and Mengaldo]{turbe2023posthoc}
Hugues Turb\'e, Mina Bjelogrlic, Christian Lovis, and Gianmarco Mengaldo.
\newblock Evaluation of post-hoc interpretability methods in time-series classification.
\newblock \emph{Nature Machine Intelligence}, 2023.

\bibitem[Wang et~al.(2021)Wang, Bau, and Zhu]{wang2021sketch}
Sheng-Yu Wang, David Bau, and Jun-Yan Zhu.
\newblock Sketch your own gan.
\newblock In \emph{CVPR}, 2021.

\bibitem[Wei et~al.(2018)Wei, Lim, Zisserman, and Freeman]{arrow-of-time}
Donglai Wei, Joseph Lim, Andrew Zisserman, and William~T. Freeman.
\newblock Learning and using the arrow of time.
\newblock In \emph{CVPR}, 2018.

\bibitem[Wei et~al.(2022)Wei, Nair, Dhurandhar, Varshney, Daly, and Singh]{wei2022interpretablesafety}
Dennis Wei, Rahul Nair, Amit Dhurandhar, Kush~R. Varshney, Elizabeth~M. Daly, and Moninder Singh.
\newblock On the safety of interpretable machine learning: A maximum deviation approach.
\newblock In \emph{NeurIPS}, 2022.

\bibitem[Xiao et~al.(2021)Xiao, Yu, Han, Zheng, and Fu]{sketchhairsalon}
Chufeng Xiao, Deng Yu, Xiaoguang Han, Youyi Zheng, and Hongbo Fu.
\newblock Sketchhairsalon: deep sketch-based hair image synthesis.
\newblock \emph{ACM TOG}, 2021.

\bibitem[Xu et~al.(2018)Xu, Huang, Yuan, Pang, Song, Xiang, Hospedales, Ma, and Guo]{xu2018sketchmate}
Peng Xu, Yongye Huang, Tongtong Yuan, Kaiyue Pang, Yi-Zhe Song, Tao Xiang, Timothy~M Hospedales, Zhanyu Ma, and Jun Guo.
\newblock Sketchmate: Deep hashing for million-scale human sketch retrieval.
\newblock In \emph{CVPR}, 2018.

\bibitem[Xu et~al.(2022)Xu, Han, Hui, Qian, and Xie]{xu2022domain}
Rui Xu, Zongyan Han, Le Hui, Jianjun Qian, and Jin Xie.
\newblock Domain disentangled generative adversarial network for zero-shot sketch-based 3d shape retrieval.
\newblock \emph{AAAI}, 2022.

\bibitem[Yan et~al.(2020)Yan, Chen, Yang, and Wang]{yan2020interactive}
Guowei Yan, Zhili Chen, Jimei Yang, and Huamin Wang.
\newblock Interactive liquid splash modeling by user sketches.
\newblock \emph{ACM TOG}, 2020.

\bibitem[Yang et~al.(2021)Yang, Shen, and Zhou]{semantic-hierarchy}
Ceyuan Yang, Yujun Shen, and Bolei Zhou.
\newblock Semantic hierarchy emerges in deep generative representations for scene synthesis.
\newblock \emph{IJCV}, 2021.

\bibitem[Yeh et~al.(2019)Yeh, Hsieh, Suggala, Inouye, and Ravikumar]{yeh2019fidelity}
Chih-Kuan Yeh, Cheng-Yu Hsieh, Arun Suggala, David~I Inouye, and Pradeep~K Ravikumar.
\newblock On the (in) fidelity and sensitivity of explanations.
\newblock In \emph{NeurIPS}, 2019.

\bibitem[Yelamarthi et~al.(2018)Yelamarthi, Reddy, Mishra, and Mittal]{yelamarthi2018zero}
Sasi~Kiran Yelamarthi, Shiva~Krishna Reddy, Ashish Mishra, and Anurag Mittal.
\newblock A zero-shot framework for sketch based image retrieval.
\newblock In \emph{ECCV}, 2018.

\bibitem[Yi et~al.(2022)Yi, Ye, Fan, Shu, Liu, Lai, and Rosin]{yi2022animating}
Ran Yi, Zipeng Ye, Ruoyu Fan, Yezhi Shu, Yong-Jin Liu, Yu-Kun Lai, and Paul~L Rosin.
\newblock Animating portrait line drawings from a single face photo and a speech signal.
\newblock In \emph{ACM SIGGRAPH}, 2022.

\bibitem[Yu et~al.(2019)Yu, Lin, Yang, Shen, Lu, and Huang]{yu2019free}
Jiahui Yu, Zhe Lin, Jimei Yang, Xiaohui Shen, Xin Lu, and Thomas~S Huang.
\newblock Free-form image inpainting with gated convolution.
\newblock In \emph{CVPR}, 2019.

\bibitem[Yu et~al.(2015)Yu, Yang, Song, Xiang, and Hospedales]{yu2015sketch}
Qian Yu, Yongxin Yang, Yi-Zhe Song, Tao Xiang, and Timothy Hospedales.
\newblock Sketch-a-net that beats humans.
\newblock In \emph{BMVC}, 2015.

\bibitem[Yu et~al.(2016)Yu, Liu, Song, Xiang, Hospedales, and Loy]{yu2016sketch}
Qian Yu, Feng Liu, Yi-Zhe Song, Tao Xiang, Timothy~M Hospedales, and Chen-Change Loy.
\newblock Sketch me that shoe.
\newblock In \emph{CVPR}, 2016.

\bibitem[Zeiler and Fergus(2014)]{zeiler2014visualising}
Matthew~D Zeiler and Rob Fergus.
\newblock Visualizing and understanding convolutional networks.
\newblock In \emph{ECCV}, 2014.

\bibitem[Zeng et~al.(2022)Zeng, Lin, and Patel]{zeng2022sketchedit}
Yu Zeng, Zhe Lin, and Vishal~M Patel.
\newblock Sketchedit: Mask-free local image manipulation with partial sketches.
\newblock In \emph{CVPR}, 2022.

\bibitem[Zhang and Wang(2019)]{zhang2019attack}
Haichao Zhang and Jianyu Wang.
\newblock Towards adversarially robust object detection.
\newblock In \emph{ICCV}, 2019.

\bibitem[Zhang et~al.(2016)Zhang, Liu, Zhang, Ren, Wang, and Cao]{zhang2016sketchnet}
Hua Zhang, Si Liu, Changqing Zhang, Wenqi Ren, Rui Wang, and Xiaochun Cao.
\newblock Sketchnet: Sketch classification with web images.
\newblock In \emph{CVPR}, 2016.

\bibitem[Zhang et~al.(2018)Zhang, Bargal, Lin, Brandt, Shen, and Sclaroff]{zhang2018top}
Jianming Zhang, Sarah~Adel Bargal, Zhe Lin, Jonathan Brandt, Xiaohui Shen, and Stan Sclaroff.
\newblock Top-down neural attention by excitation backprop.
\newblock \emph{IJCV}, 2018.

\bibitem[Zhang et~al.(2023)Zhang, Rao, and Agarwala]{controlnet}
Lvmin Zhang, Anyi Rao, and Maneesh Agarwala.
\newblock Adding conditional control to text-to-image diffusion models.
\newblock In \emph{ICCV}, 2023.

\bibitem[Zhang et~al.(2021)Zhang, Guo, and Gu]{zhang2021sketch2model}
Song-Hai Zhang, Yuan-Chen Guo, and Qing-Wen Gu.
\newblock Sketch2model: View-aware 3d modeling from single free-hand sketches.
\newblock In \emph{CVPR}, 2021.

\bibitem[Zhou et~al.(2016)Zhou, Khosla, Lapedriza, Oliva, and Torralba]{zhou2016learning}
Bolei Zhou, Aditya Khosla, Agata Lapedriza, Aude Oliva, and Antonio Torralba.
\newblock Learning deep features for discriminative localization.
\newblock In \emph{CVPR}, 2016.

\bibitem[Zhou et~al.(2020)Zhou, Chen, Liao, Zhang, Chen, Dong, Liu, Hua, and Yu]{zhou2020LGGAN}
Hang Zhou, Dongdong Chen, Jing Liao, Weiming Zhang, Kejiang Chen, Xiaoyi Dong, Kunlin Liu, Gang Hua, and Nenghai Yu.
\newblock Lg-gan: Label guided adversarial network for flexible targeted attack of point cloud-based deep networks.
\newblock In \emph{CVPR}, 2020.

\bibitem[Zou et~al.(2018)Zou, Yu, Du, Mo, Song, Xiang, Gao, Chen, and Zhang]{zou2018sketchyscene}
Changqing Zou, Qian Yu, Ruofei Du, Haoran Mo, Yi-Zhe Song, Tao Xiang, Chengying Gao, Baoquan Chen, and Hao Zhang.
\newblock Sketchyscene: Richly-annotated scene sketches.
\newblock In \emph{ECCV}, 2018.

\end{thebibliography}
}

\clearpage

\appendix

\vspace{1cm}
\clearpage

\setcounter{page}{1}

\maketitlesupplementary

\section{Differentiable Rasterisation for P-SLA}

\begin{figure}[htbp]
    \vspace{-3mm}
    \includegraphics[width=\linewidth]{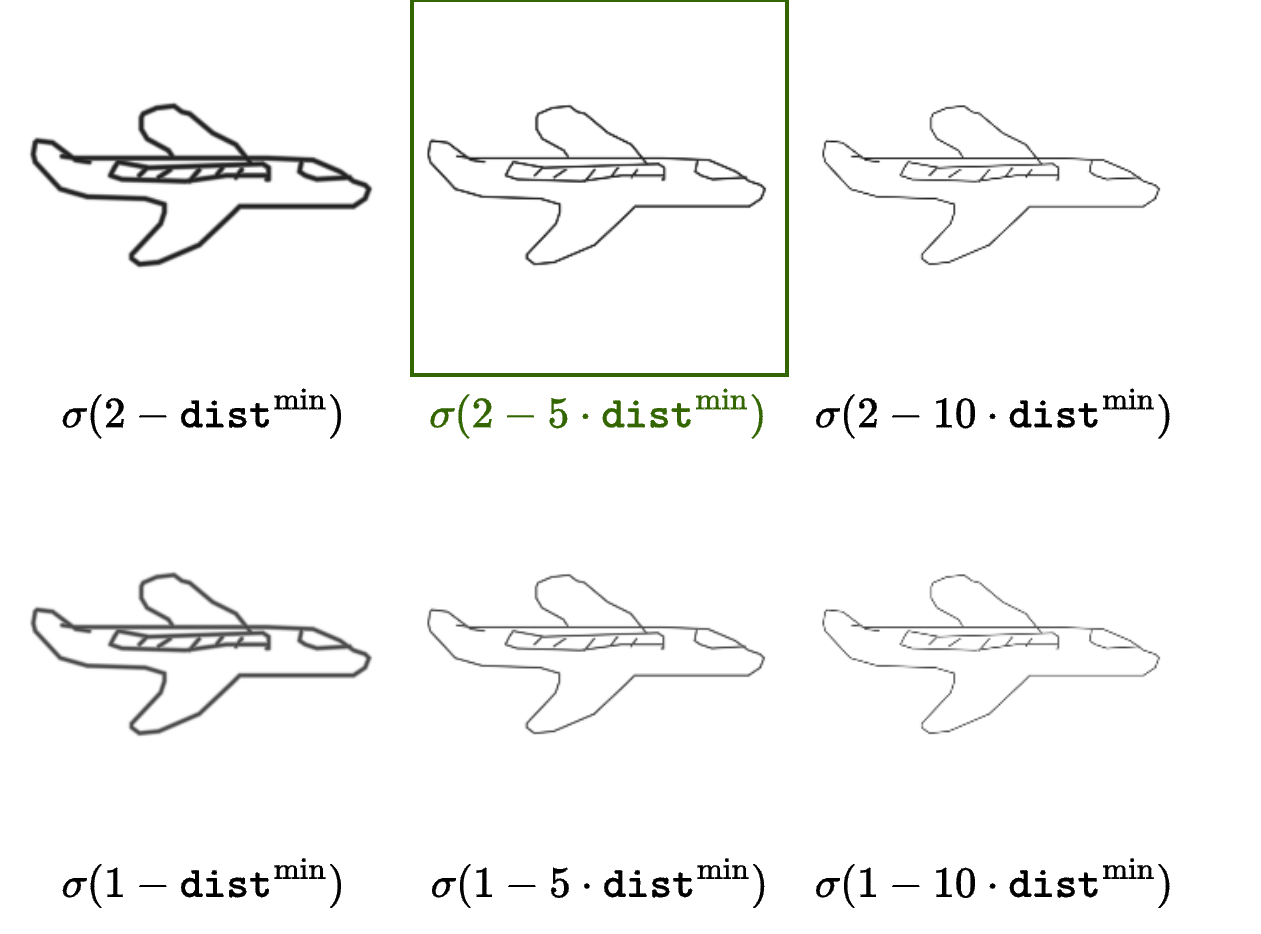}
    \vspace{-3mm}
    \caption{ Modulating stroke thickness by a soft-threshold on the minimum distance $\texttt{dist}^{\text{min}}$ of each pixel from any stroke.\label{fig:dist_supp}}
    \vspace{-2mm}
\end{figure}

We differentiably render vector sketches $\mathrm{V} \in \mathbb{R}^{T \times 5}$ (differentiably convert vector \(\rightarrow\) raster) by (i) calculating the minimum distance of each pixel in a blank canvas ($\mathrm{X} \in \mathbb{R}^{H \times W \times 3}$) from any stroke in $\mathrm{V}$, and (ii) colouring all pixels by their distance, controlled with threshold hyperparameters. Essentially, these hyperparameters control how thick the rendered strokes are ({\cref{fig:dist_supp}}), by regulating {pixels' colour} based on how far they sit from the {stroke}.

\begin{figure}[htbp]
    \vspace{-3mm}
    \includegraphics[width=\linewidth]{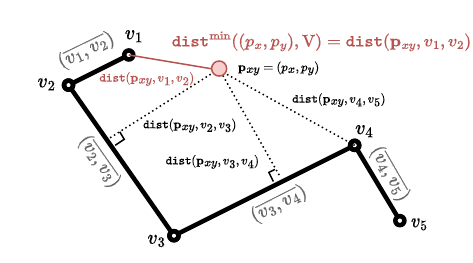}
    \vspace{-3mm}
    \caption{Calculation of minimum distance of pixel-coordinate \(\mathbf{p}_{xy}=(p_x,p_y)\) from vector sketch $\mathrm{V}\in \mathbb{R}^{T \times 5}$ where $T=5$  \label{fig:vsa_supp}}
    \vspace{-2mm}
\end{figure}

We simplify the first problem by calculating the distance of each pixel ($p_x,p_y$) from \textit{linear segments} of vector strokes $(\overline{v_{t-1},v_t})$ \textit{i.e.}, consecutive points in $\mathrm{V}$ (\cref{fig:vsa_supp}). Next, we find the minimum of these distances from all such line segments to get the minimum distance from all strokes in $\mathrm{V}$. 
Now, the distance of pixel coordinate $\mathbf{p}_{xy}=(p_x,p_y)$ from line segment $(\overline{v_{t-1},v_t})$ is:
\vspace{-0.1cm}
\begin{equation}
\footnotesize
\begin{split}
     \texttt{dist}(\mathbf{p}_{xy},v_{t-1}, v_t) = \hspace{5cm} &  \\
    \begin{cases}
      |(\overline{\mathbf{p}_{xy},v_t})|, & \text{if}\ \angle \mathbf{p}_{xy}\; v_t\; v_{t-1} > \frac{\pi}{2} \\
      |(\overline{\mathbf{p}_{xy},v_{t-1}})|, & \text{if}\ \angle \mathbf{p}_{xy}\; v_{t-1}\; v_{t} > \frac{\pi}{2} \\
     |(\overline{\mathbf{p}_{xy},v_t})\times (\overline{v_t,v_{t-1}})|\div|(\overline{v_t,v_{t-1}})|, & \text{otherwise}\\
    \end{cases}
\end{split}
\end{equation}
where \textit{(i)} $|(\overline{A,B})|$ denotes the euclidean distance between coordinate points $A$ and $B$, \textit{(ii)} $(\overline{A,B}) \times (\overline{C,D})$ denotes cross product between vectors $(\overline{A,B})$ and $(\overline{C,D})$, and \textit{(iii)} $\angle ABC$ denotes the angle formed by coordinates $A$, $B$, and $C$ with $B$ as the vertex. 

\noindent This equation is implemented as the distance of pixel coordinate $\mathbf{p}_{xy}=(p_x,p_y)$ from line segment $(\overline{v_{t-1},v_{t}})$ in \cref{alg:compute-dist}. For each coordinate $\mathbf{p}_{xy}$, we compute the minimum distance as \(\displaystyle \texttt{dist}^\text{min}(\mathbf{p}_{xy},\mathrm{V}) = \min_{t=2, \dots, T} (\texttt{dist}(\mathbf{p}_{xy},v_{t-1},v_t))\) over all vector points in $\mathrm{V}$ to check whether this distance is under a given threshold. However, not all consecutive points in $\mathrm{V}$ are connected. At the end of one stroke (say, $v_{t-1}$) and the beginning of the next ($v_t$), the pen is lifted and moved to the new coordinate without drawing on the canvas. This motion is indicated with a pen-up state ($q^{1}_{t-1}=0$). To prevent joining these points in $(\overline{v_{t-1},v_{t}})$ in the final raster sketch, we exclude them from the minimum distance calculation by offsetting the distance value $\texttt{dist}(\mathbf{p}_{xy},v_{t-1},v_t)$ with a large number ($10^6$). Then, we calculate the minimum distance as : 
\vspace{-0.2cm}
\begin{equation}
\begin{split}
\hspace{-0.3cm}\texttt{dist}^\text{min} (\mathbf{p}_{xy}, \mathrm{V}) =  \min_{t=2, \dots, T}\Big(&\texttt{dist}(\mathbf{p}_{xy},v_{t-1},v_t)  \\
& \hspace{0.7cm} + (1-q^{1}_{t-1})10^6\Big)    
\end{split}
\end{equation}

\vspace{-0.2cm}
\noindent {The colour ($1$$\to$black, $0$$\to$white) of a pixel $\mathrm{X}(p_x,p_y)$ is determined by the minimum distance $\texttt{dist}^\text{min}((p_x,p_y),\mathrm{V})$ where $\mathbf{p}_{xy}=(p_x,p_y)$, as:}
\begin{equation}
    \mathrm{X} (p_x,p_y) = \sigma(2-5 \cdot \texttt{dist}^\text{min}((p_x,p_y),\mathrm{V}))
\end{equation}
{where \(\sigma\) represents the sigmoid function and \(\mathrm{X} \in \mathbb{R}^{H \times W \times 3}\) represents the raster sketch image. Here, $\sigma$ acts like a \textit{soft-threshold} to convert $\texttt{dist}^\text{min}$ to either $\sim$$1$ (black colour pixel) or $\sim$$0$ (white colour pixel).
We control stroke thickness in \cref{fig:dist_supp} with the threshold hyperparameters empirically set to $2$ and $5$.}

\section{Additional Results on Adversarial attacks}

\includegraphics[width=\linewidth]{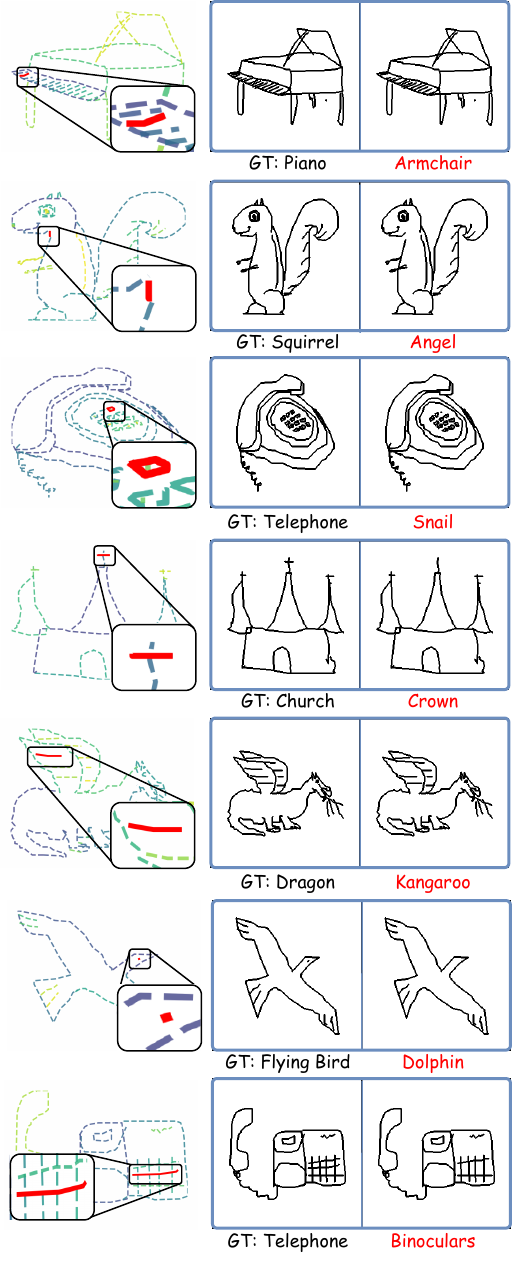}

\section{Future Directions}
We introduce fine-grained attributions in sketch-based networks with a plug-and-play explainability toolbox. While we use vanilla gradient-based attributions as a design choice for simplicity, strokes can be attributed with more intricate attribution algorithms \cite{kapishnikov2019xrai, kapishnikov2021guided, liu2017deep, smilkov2017smoothgrad} in future works. Specifically, SLA and P-SLA can be paired with any pixel-attribution algorithm (like Guided Integrated Gradients \cite{kapishnikov2021guided}), where, from attributions of all pixels only those for pixels containing the stroke $(\mathrm{X}(p_x,p_y))$ can be selected. We emphasise that selecting an optimal attribution algorithm is out-of-scope of this paper, as here we primarily demonstrate the applicability of fine-grained attributions in downstream sketch tasks.

\section{Algorithm to compute \texttt{dist($\cdot$)}}
\label{sec: compute-dist}

\begin{algorithm}[h]
    \caption{Compute \texttt{dist}($\cdot$)}\label{alg:compute-dist}
    \SetKwFunction{FMain}{\texttt{dist}}
    \SetKwProg{Fn}{Function}{:}{}
    \Fn{\FMain{$p_{x}, p_{y}, v_{t-1}, v_{t}$}}{
        $(x_{t-1}, y_{t-1}, q_{t-1}^{1}, q_{t-1}^{2}, q_{t-1}^{3}) \gets v_{t-1} $ \;
        $(x_{t}, y_{t}, q_{t}^{1}, q_{t}^{2}, q_{t}^{3}) \gets v_{t} $ \;
        $\delta x \gets x_{t} - x_{t-1}$ \;
        $\delta y \gets y_{t} - y_{t-1}$ \;
        $\text{norm} \gets px \cdot px + py \cdot py$ \;
        $u' \gets ( (p_{x} - x_{t-1}) \cdot \delta x + (p_{y} - y_{t-1}) \cdot \delta y$ \;
        $u \gets u' / \text{norm}$ \;
        \If{ $u > 1$}{
            $u \gets 1$ \;
        }
        \ElseIf{ $u < 0$}{
            $u \gets 0$
        }
        $x \gets x_{t-1} + u \cdot \delta x$ \;
        $y \gets y_{t-1} + u \cdot \delta y$ \;
        $\Delta x \gets x - p_{x}$ \;
        $\Delta y \gets y - p_{y}$ \;
        \textbf{return} $(\Delta x \cdot \Delta x + \Delta y \cdot \Delta y)^{1/2}$
    }
\end{algorithm}

\end{document}